\documentclass[journal]{IEEEtran}

\usepackage{float} \restylefloat{figure} 
\usepackage{xcolor}
\usepackage{color} 

\usepackage{graphicx}
\graphicspath{{eps/}}
\usepackage{multirow}
\usepackage{threeparttable}
\usepackage{gensymb}
\usepackage{rotating}
\usepackage{lscape}
\usepackage{bm} 
\usepackage{cite}
\usepackage{url}
\usepackage[small]{caption}
\usepackage{subfigure} 
\usepackage{array} 
\usepackage{booktabs}
\usepackage{amsmath}   
\usepackage{amsfonts}
\usepackage{afterpage}
\usepackage{afterpage}
\usepackage{epstopdf}
\long\def\symbolfootnote[#1]#2{\begingroup\def\thefootnote{\fnsymbol{footnote}}
\footnote[#1]{#2}\endgroup}

\begin{document}
\title{VLSI Extreme Learning Machine: A Design Space Exploration}
\author{Enyi Yao, \IEEEmembership{Student Member,~IEEE} and Arindam Basu, \IEEEmembership{Member,~IEEE}
\thanks{The authors are with the School of Electrical and Electronic Engineering, Nanyang Technological University, Singapore. (email: eyao1@e.ntu.edu.sg,  arindam.basu@ntu.edu.sg)}}

\maketitle
\begin{abstract}
In this paper, we describe a compact low-power, high performance hardware implementation of the extreme learning machine (ELM) for machine learning applications. Mismatch in current mirrors are used to perform the vector-matrix multiplication that forms the first stage of this classifier and is the most computationally intensive. Both regression and classification (on UCI data sets) are demonstrated and a design space trade-off between speed, power and accuracy is explored. Our results indicate that for a wide set of problems, $\sigma V_T$ in the range of $15-25$mV gives optimal results. An input weight matrix rotation method to extend the input dimension and hidden layer size beyond the physical limits imposed by the chip is also described. This allows us to overcome a major limit imposed on most hardware machine learners. The chip is implemented in a $0.35 \mu$m CMOS process and occupies a die area of  around 5 mm $\times$ 5 mm. Operating from a $1$ V power supply, it achieves an energy efficiency of $0.47$ pJ/MAC at a classification rate of $31.6$ kHz.

\end{abstract}

\begin{IEEEkeywords}
Extreme Learning Machine, Classifier, Machine Learning, Low Power, Neural Networks
\end{IEEEkeywords}

\section{Introduction}
In general, it is difficult to achieve high accuracy in pure analog signal processing modules due to several reasons, a major one being device mismatch \cite{kinget-mismatch}. The effect of mismatch on traditional circuits like differential amplifiers and current mirrors is well documented\cite{razavi-cmos}. It has also been shown that for MOS based circuits, the extra power dissipation needed to overcome effects of mismatch can be an order of magnitude higher than the limit imposed by thermal noise\cite{kinget-mismatch}. With transistor dimensions reducing over the years, variance in properties of transistors, notably the threshold voltage, has kept on increasing making it difficult to rely on conventional simulations ignoring statistical variations. The problem is particularly exacerbated for neuromorphic designs\cite{giacomo_mismatch2010}, where transistors are typically biased in the sub-threshold region \cite{giacomo_spikestdp,arthur_gamma,nullcline_neu_my} of operation (to glean maximal efficiencies in energy per operation) since device currents are exponentially related to threshold voltages thus amplifying its variations as well. For example, it is shown in \cite{bernabe-minidac} that an array of $5-bit$ DACs in $0.35\mu$m CMOS process used as tunable weights only provide an effective number of bits of $1.1$ due to mismatch. In general, there has been an approach to compensate for mismatch either through floating-gates\cite{brink_learningfg} or by storing calibration coefficients off-chip in the form of connection probabilities\cite{giacomo_mismatch2010}. Digital calibration can be used to compensate for these effects on-chip\cite{bernabe-minidac} as well. However, they lead to huge area overheads due to the requirement of extra transistors for calibration and storage of digital bits\cite{sun-biocas}. Sometimes, it is claimed that learning can compensate for mismatch and has been demonstrated in specific cases\cite{pfeil-stdp,murray-mismatch}--but the claim needs to be further quantified using standard datasets since mismatch will exist in the learning circuits as well.

The ELM algorithm is popular in the machine learning community due to its fast training speed and has been shown to produce similar or better performance compared to support vector machines (SVM)\cite{huang_elm_kernel}. A closely related method (termed Neural Engineering Framework) has also been used to generate large scale models of cognitive systems\cite{science-eliasmith}. ELM based methods have been used classify spike time based patterns recently \cite{skim} and online learning algorithms for ELM have been proposed\cite{elm-online}. Clearly there is a need to develop hardware implementations of the same. In this paper we present a circuit that `utilizes' mismatch to do effective computation in the first layer of a two layer spiking neural network implementation of ELM. This approach can be used in other algorithms like liquid state machine (LSM) or echo state networks (ESN) (sometimes referred to as reservoir computing), since they require random projections of the input as well.  We have earlier proposed the idea of using spiking neurons for implementing ELM\cite{basu_shuo_elm} and described the advantages of such an architecture over standard digital implementations\cite{elm-biocas}. It should be noted that this method only exploits spiking neurons for ease of hardware implementation and does not use any spike based learning rules to perform the learning of the second stage. The major hardware benefits are the use of low-power analog circuits for the reservoir and simple digital circuits for the second stage. We demonstrated the first VLSI implementation of this principle in \cite{chenyi_iscas2015} where it was used for decoding motor intentions for implantable brain-machine interfaces. In this paper, we present a different chip utilizing the same core circuit as \cite{chenyi_iscas2015} but operating on $10$ bit digital inputs instead of spikes. Instead of a specific application, this paper presents an entire design space trade-off between speed, power and accuracy. Finally, we present a method and associated circuits to virtually expand the input and output dimensions of the chip beyond the physically implemented $128$ channels. We show results of applying inputs from standard machine learning data bases such as \cite{uci}.

In the next section, we present details of the ELM algorithm and training methods. Section \ref{sec:architecture} describes the VLSI architecture of the chip and details of the sub-circuits. The trade-offs between noise, speed and energy dissipation of this architecture are presented in Section \ref{sec:energy}. An important limitation of hardware machine learners is limited input and output dimensions. In Section \ref{sec:dimension}, we present a method to virtually expand the dimensions beyond the physical number of channels on the chip. Measurement results are presented in Section \ref{sec:results} and finally we conclude in the last section.

\begin{figure}[t]
	\centerline{
		\includegraphics[width=0.35\textwidth]{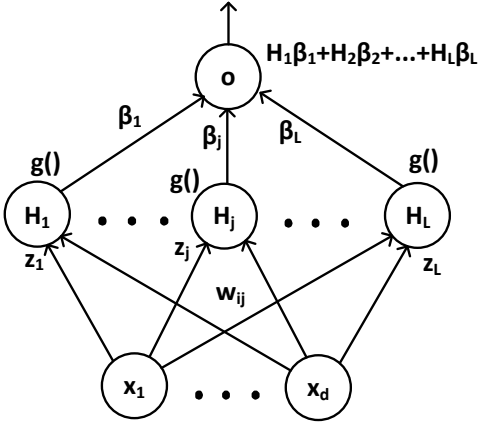} }
	\caption{The architecture of ELM algorithm. $d$ is the dimension of the input data and $L$ denotes the number of hidden layer neurons.}\label{fig:algorithm}
\end{figure}

\section{ELM Theory}
\label{sec:theory}
In this section, we will present a brief description of the ELM algorithm and refer the readers to \cite{huang_elm_kernel,huang_elm_survey} for details. As illustrated in Fig. \ref{fig:algorithm}, the ELM algorithm is applicable to a two layer neural feed-forward network with L hidden neurons having an activation function $g:R\rightarrow R$. Without loss of generality, we consider a scalar output in this case though the method can be easily extended to multiple outputs by considering each output one by one\cite{huang_elm}. The output of the network $o$ is given by:
\begin{equation}
\label{eq:elm_eqn}
\begin{split}
o &=\sum _i^L \beta_i H_i=\sum _i^L \beta_i g\left(z_i\right)\\
 &=\sum _i^L \beta_i g(\mathbf{w_i^Tx}+b_i), \mathbf{w_i,x}\epsilon R^d, \beta_i, b_i\epsilon R,\\
\end{split}
\end{equation}
where $\beta$ denote the output weights, $z_i$ and $H_i$ are the input and output of the i-th hidden layer neuron. $\bf{w}_i$ denotes the input weight and $b_i$ is the bias for the i-th neuron. In general, a sigmoidal form of $g()$ is assumed though other functions have also been used. Compared to traditional back propagation learning rule that modifies all the weights, the ELM allows $\bf{w}_i$ and $b_i$ to be random numbers drawn from any continuous distribution while only the output weights, $\beta_i$ needs to be tuned based on the training data $T$. For $N$ samples $(\textbf{x}_k,t_k)$, the hidden layer output matrix $\textbf{H}$ is defined as:
\begin{equation}
\label{eq:elm_hidden}
H=\begin{bmatrix}
g(\mathbf{w_1^Tx_1}+b_1)&...&g(\mathbf{w_L^Tx_1}+b_L)\\
.&....&.\\
.&....&.\\
g(\mathbf{w_1^Tx_N}+b_1)&...&g(\mathbf{w_L^Tx_N}+b_L)
\end{bmatrix}
\end{equation}
The desired output weights, $\widehat{\bf{\beta}}$ are then the solution of the following optimization problem:
\begin{equation}
\label{eq:elm_op}
\text{Minimize}_\beta: \parallel\mathbf{H}\beta-\mathbf{T}\parallel ^2,
\end{equation}
where $\beta=[\beta_1..\beta_L]$ and $T=[t_1..t_N]$.
The ELM algorithm proves that the optimal solution $\bf{\widehat{\beta}}$ is given by $\bf{\widehat{\beta}}=\bf{H^\dagger T}$ where $\bf{H^\dagger}$ denotes the Moore Penrose generalized inverse of a matrix\cite{huang_elm_kernel}. The huge benefit of this method is that it removes the need for iterative tuning and gives a simple formula to calculate the weights. The orthogonal projection method can be efficiently used to find $\bf{H^\dagger}$ as $(\bf{H}^T\bf{H})^{-1}\bf{H}^T$ if $\bf{H}^T\bf{H}$ is non-singular or as $\bf{H}^T(\bf{H}\bf{H}^T)^{-1}$ if $\bf{H}\bf{H}^T$ is nonsingular. Further, using concepts from ridge regression theory \cite{1970_HoerlKennard}, a small constant $I/C$ is often added to the diagonal of $\bf{H}^T\bf{H}$ or $\bf{H}\bf{H}^T$ of the Moore-Penrose generalized inverse $\textbf{H}$--the resultant resolution is stabler and tends to have better generalization performance. The value of $C$ is typically optimized as a hyperparameter using cross-validation techniques. 

\begin{figure}[h]
	\centering
	
	\includegraphics[width=0.475\textwidth]{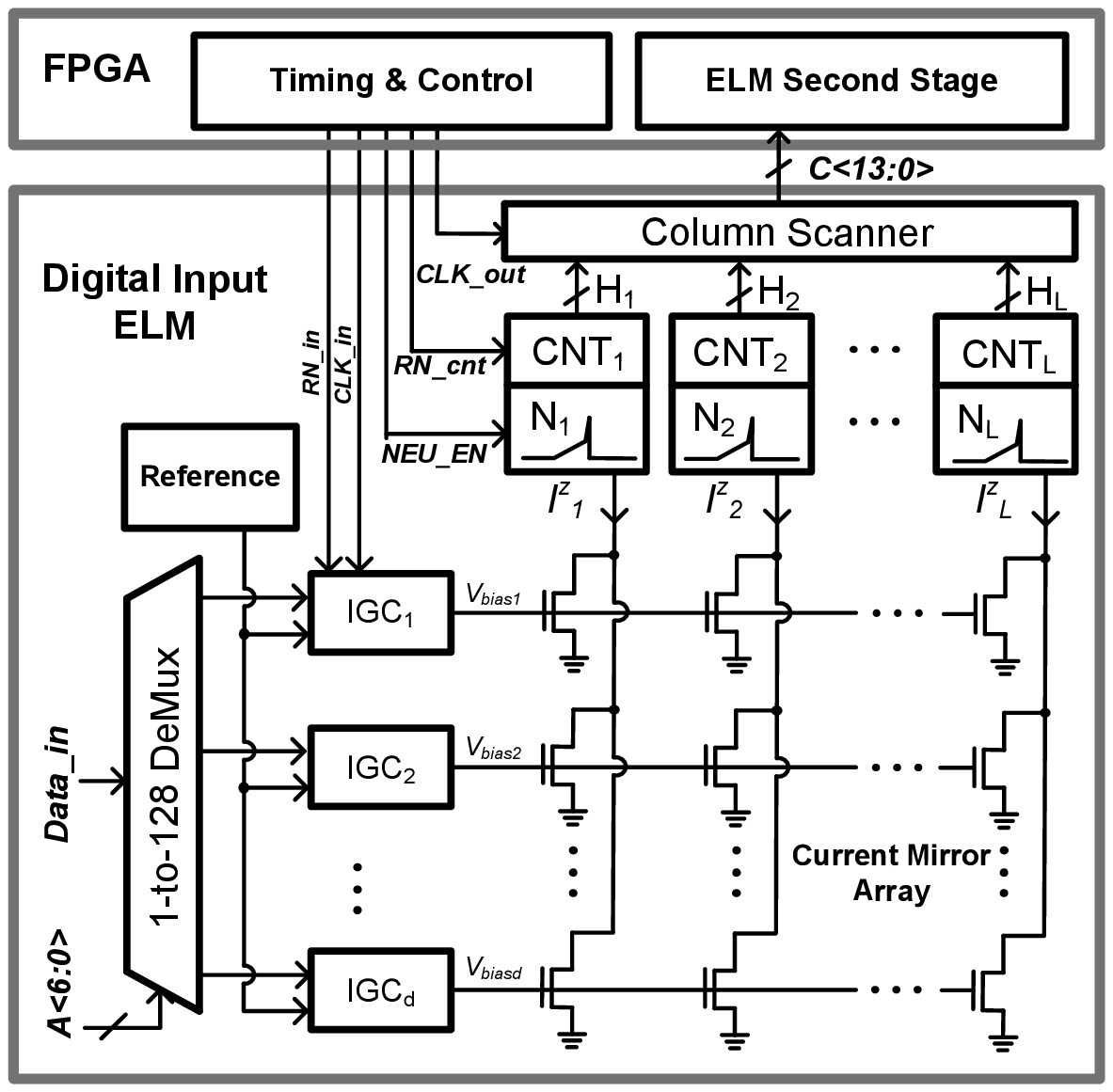} \\(a)\\
	\includegraphics[width=0.4\textwidth]{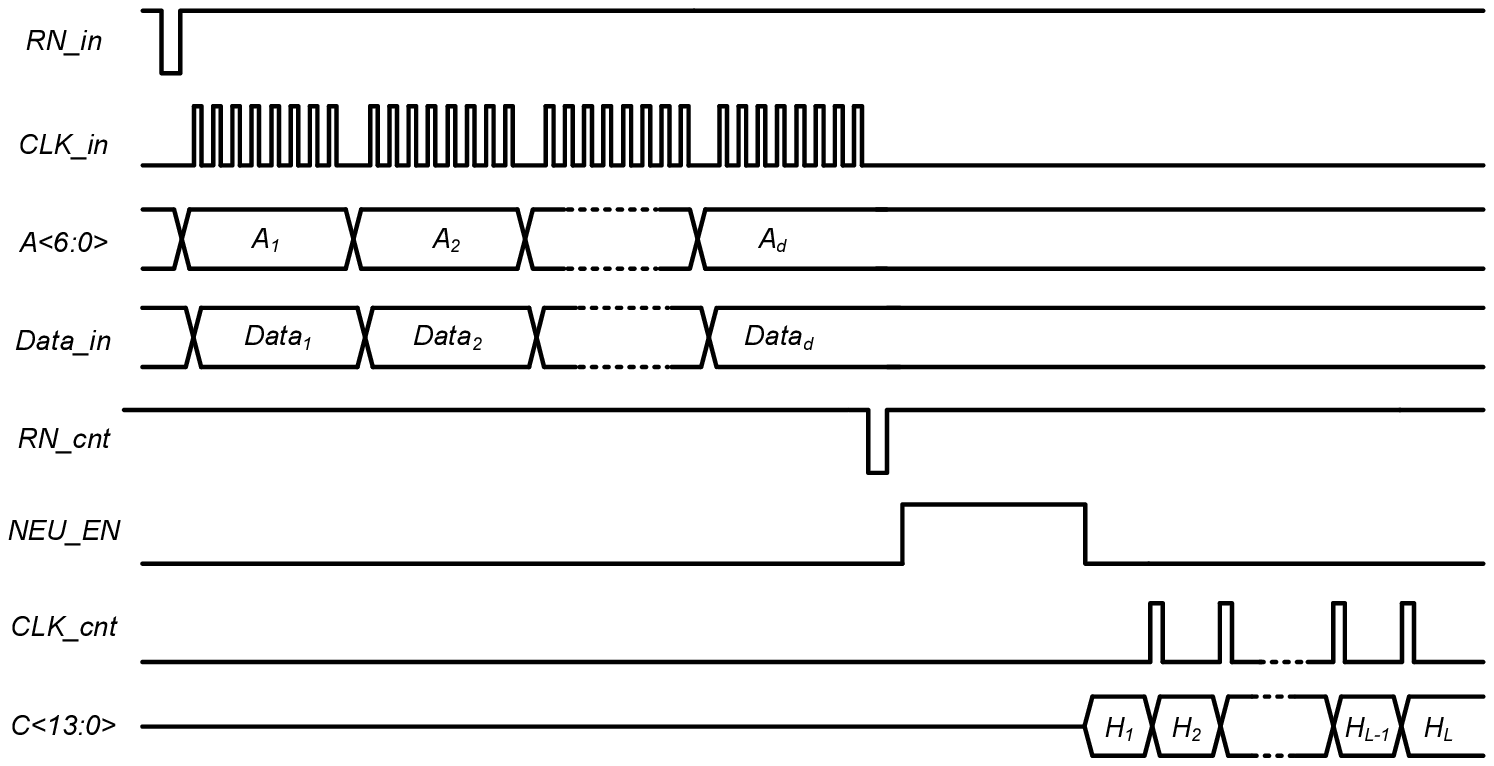} \\(b)
	\caption{(a) System architecture of the mixed signal integrated circuit that implements the first stage of ELM; the second stage is implemented in digital domain. The digital input data is converted to current by the IGC and then multiplied by random weights $w_{ij}$ in the current mirror array. The current is converted to digital domain by the combination of a spiking neuron and a counter. (b) Timing diagram of the ELM system where $RN\_in$ is a global reset, $Data\_in$, $A$ and $CLK\_in$ are SPI control signals to transfer input data to the IC. $NEU\_EN$ enables the neuron to produce spikes while $CLK\_cnt$ is used to read out the counter values $C$ one by one.}\label{fig:architecture}
\end{figure}

\section{System Architecture}
\label{sec:architecture}
The architecture of the proposed mixed signal classifier that exploits analog computing for the $d\times L$ random weights of the input layer is shown in Fig. \ref{fig:architecture}(a). The corresponding timing diagram is shown in Fig. \ref{fig:architecture} (b). The input data ($Data\underline{\ }in$) will be fed to the particular channel in the system serially through a $1$ to $128$ demultiplexor according to the corresponding address $A<6:0>$ through a serial peripheral interface (SPI). The number of bits (NOB) of $Data\underline{\ }in$ for each channel is $b_{in} = 10$. Input data will be stored in shift registers first for the configuration of the current-mode digital-to-analog convertor (DAC) in the input-generation-circuit (IGC). The function of IGC is to generate an analog DC current according to the input data which will be copied to every column using a current mirror. Multiplied by the random weights generated in the current mirror array, the current in one column will be summed according to Kirchoff's current law (KCL) and flow into a hidden layer neuron. This current is denoted as $I^z_i$ for the i-th neuron in Fig. \ref{fig:architecture}(a) and is analogous to the variable $z_i$ in Fig. \ref{fig:algorithm}. Spiking oscillations with different frequency will be generated by the neuron according to their own input currents which is counted by an asynchronous counter  forming a row of the matrix $\mathbf{H}$. Through a column scanner, these hidden layer outputs can be transferred to the FPGA to first get the output weight $\beta$ during training and later for the second stage computation of ELM during regular operation. Other timing and control signals will also be provided by the FPGA as shown in Fig. \ref{fig:architecture}(b). Next, we describe the operation of each block.

\begin{figure}[t]
	\centerline{
		\includegraphics[width=0.475\textwidth]{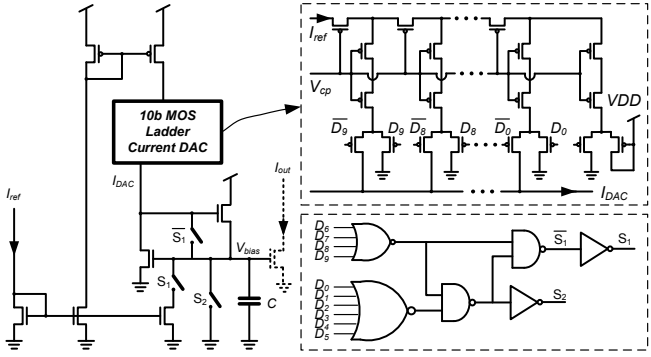} }
	\caption{Schematic of input generation circuit (IGC) for one channel. A reference current is split according to the $10$ bits of input data to create $I_{DAC}$. The capacitor $C$ ensures sufficient SNR when the current is mirrored to the $L$ columns. An active current mirror is enabled to allow fast settling when $I_{DAC}$ is small.}\label{fig:igc}
\end{figure}
\subsection{Input Generation Circuit (IGC)}
\label{sec:igc}
Figure \ref{fig:igc} shows the schematic of the input generation circuit for each dimension of input. The reference block provides a fixed master biasing current $I_{ref}$ that acts as the reference current of the current DAC as well as the biasing for the active current mirror. The input data $Data\underline{\ }in$ is applied to configure a $b_{in}=10$ bits MOS based current splitting DAC to generate a corresponding analog current\cite{tobi_bias}.  The output current of this DAC is given by:
\begin{equation}
\label{i_dac}
I_{DAC}=\left(2^{-1}D_9+2^{-2}D_8+\cdots+2^{-9}D_1+2^{-10}D_0\right)I_{ref}.
\end{equation} 
$I_{DAC}$ is multiplied with the input weights by current mirroring operation as described later. A capacitor $C=0.4$pF is also added at the gate of the current mirror array for each row to improve noise performance and achieve the desired resolution of $8$ bits in the multiplication--this will be discussed in the later section. In the conventional current mirror, bandwidth is in proportion to the input current. If $Data\underline{\ }in$ is too small, input currents are also small and hence the settling time of the current mirror (defined as time taken to settle to within $5\%$ of the final value) might be too large. To solve this problem, an active current mirror is added to complement the conventional mirror. Switch $S1$ is closed to turn on the active current mirror if all of the $4$ MSBs are zero. This ensures that the capacitor $C$ is charged by the large bias current and not the small input currents. When all the bits of $Data\_in$ are $0$, switch $S2$ is closed to pull $V_{bias}$ to ground and shut off the current mirrors in that row. The logical signals to control $S1$ and $S2$ are given by:
\begin{align}
\label{s1}
S_1&=\overline{D_6+D_7+D_8+D_9}.\notag \\
S_2&=\overline{D_0+D_1+\cdots+D_8+D_9}.
\end{align}
 where $D_i$ are the bits of $Data\_in$.
 
 \begin{figure}[h]
 	\centering 	
 	\includegraphics[width=0.45\textwidth]{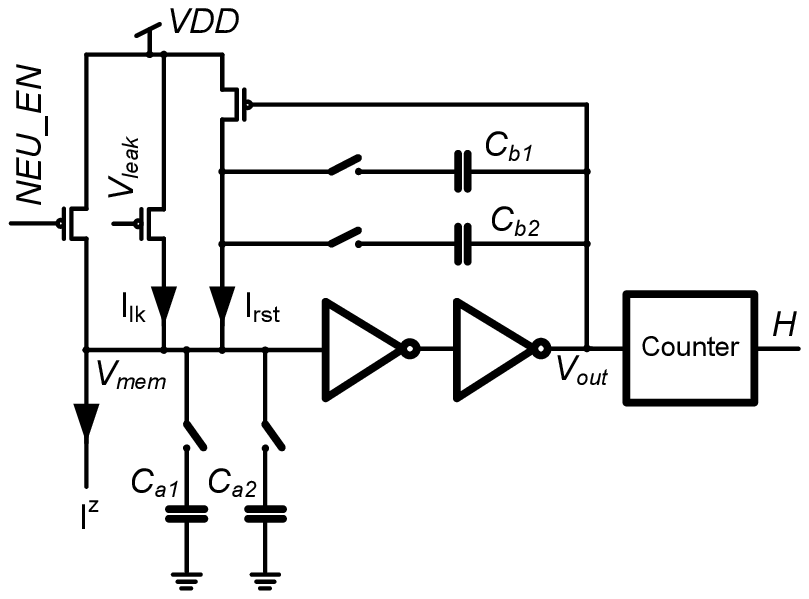} \\(a)\\
 	\includegraphics[width=0.25\textwidth]{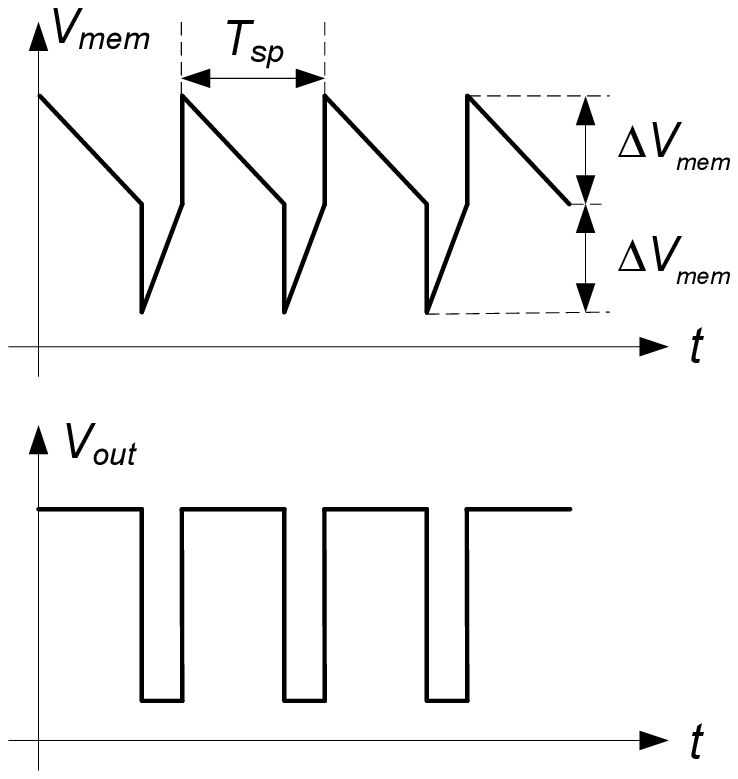} \\(b)
 	\caption{(a) Schematic of the neuronal oscillator circuit followed by an asynchronous counter. The neuron is enabled when control signal $NEU\_EN$ is high. The capacitors can be digitally reconfigured and have the following values: $C_{a1}=100$fF, $C_{a2}=200$fF, $C_{b1}=50$fF, $C_{b2}=100$fF. (b) Oscillation waveforms at different nodes of the neuron circuit.}\label{fig:vmem}
 \end{figure}
%

\begin{figure}[t]
	\centering
	\begin{minipage}{0.485\textwidth}
		\centering
		\subfigure[]{
			\label{subfig:transfer_func_a}
			\includegraphics[width=0.475\textwidth]{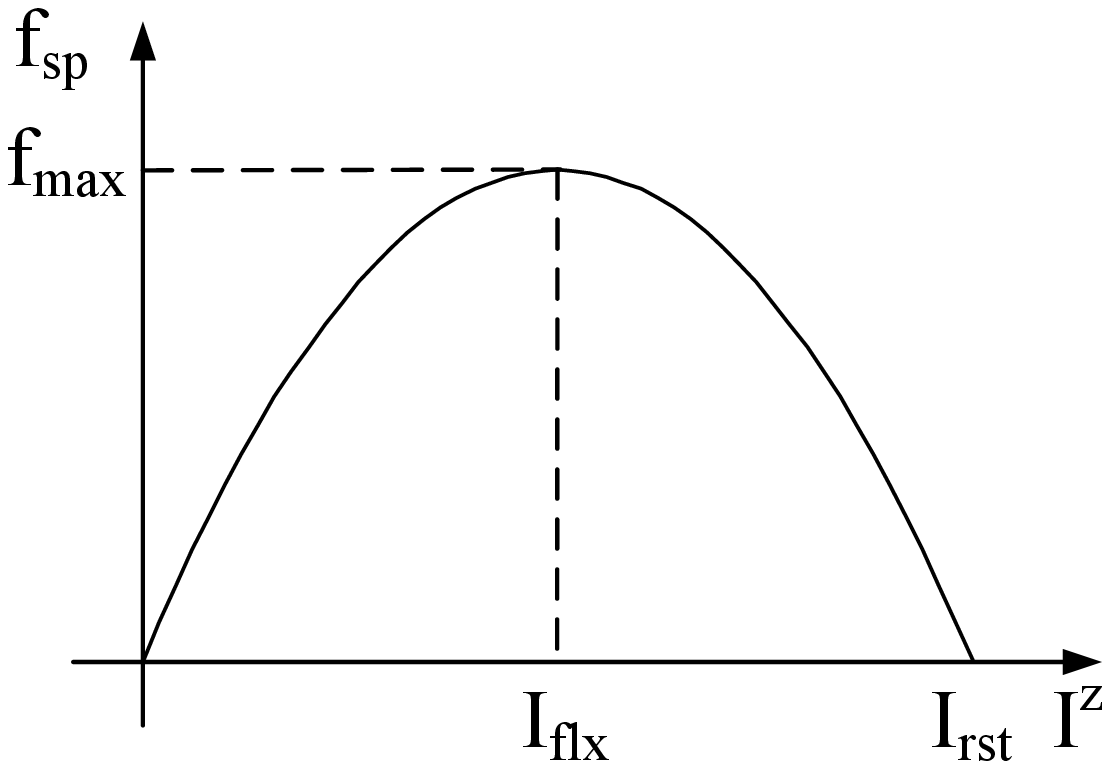}}
		\centering
		\subfigure[]{
			\label{subfig:transfer_func_b}
			\includegraphics[width=0.475\textwidth]{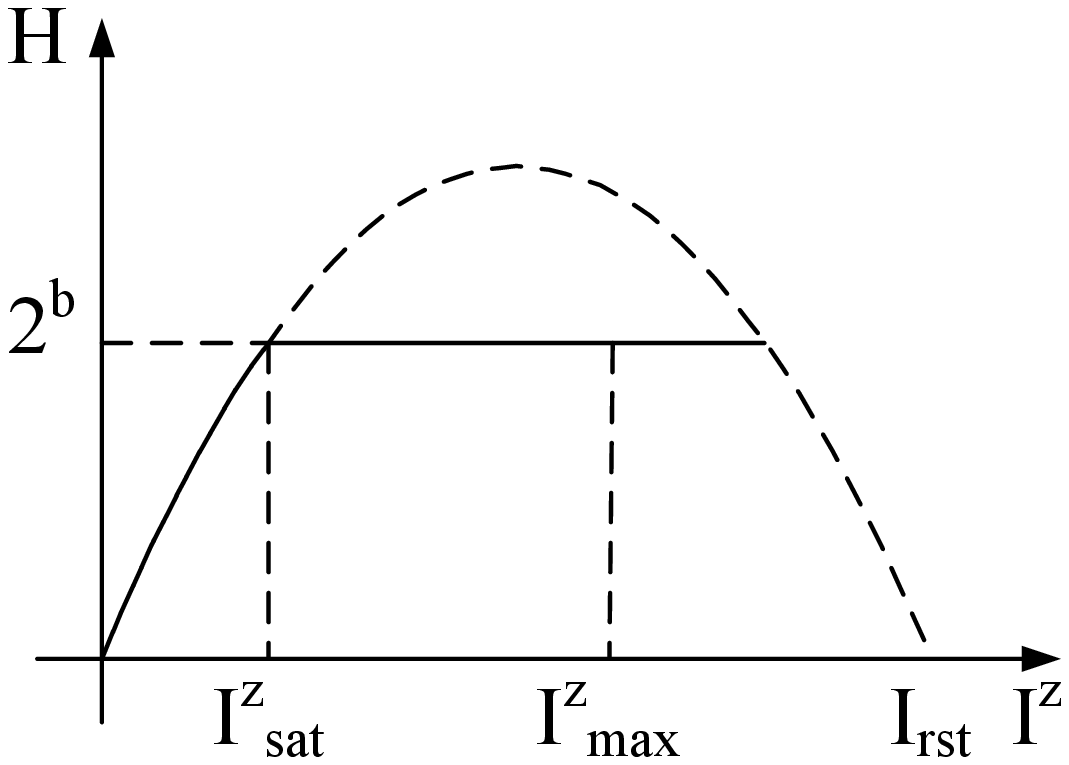}}
		\centering
		\caption{(a) Neuron spiking frequency initially increases with the increase of the input current $I^z$ till $I^z=I_{flx}$. It then reduces and becomes zero finally when $I^z=I_{rst}$. (b) The transfer function (solid line) of the neuron with input $I^z$ and output $H$ can be saturated at a pre-defined value of $2^b$ by stopping the counter.  }\label{fig:transfer_func}
	\end{minipage}
\end{figure}

 \begin{figure}[h]
 	\centering 	
 	\includegraphics[width=0.35\textwidth]{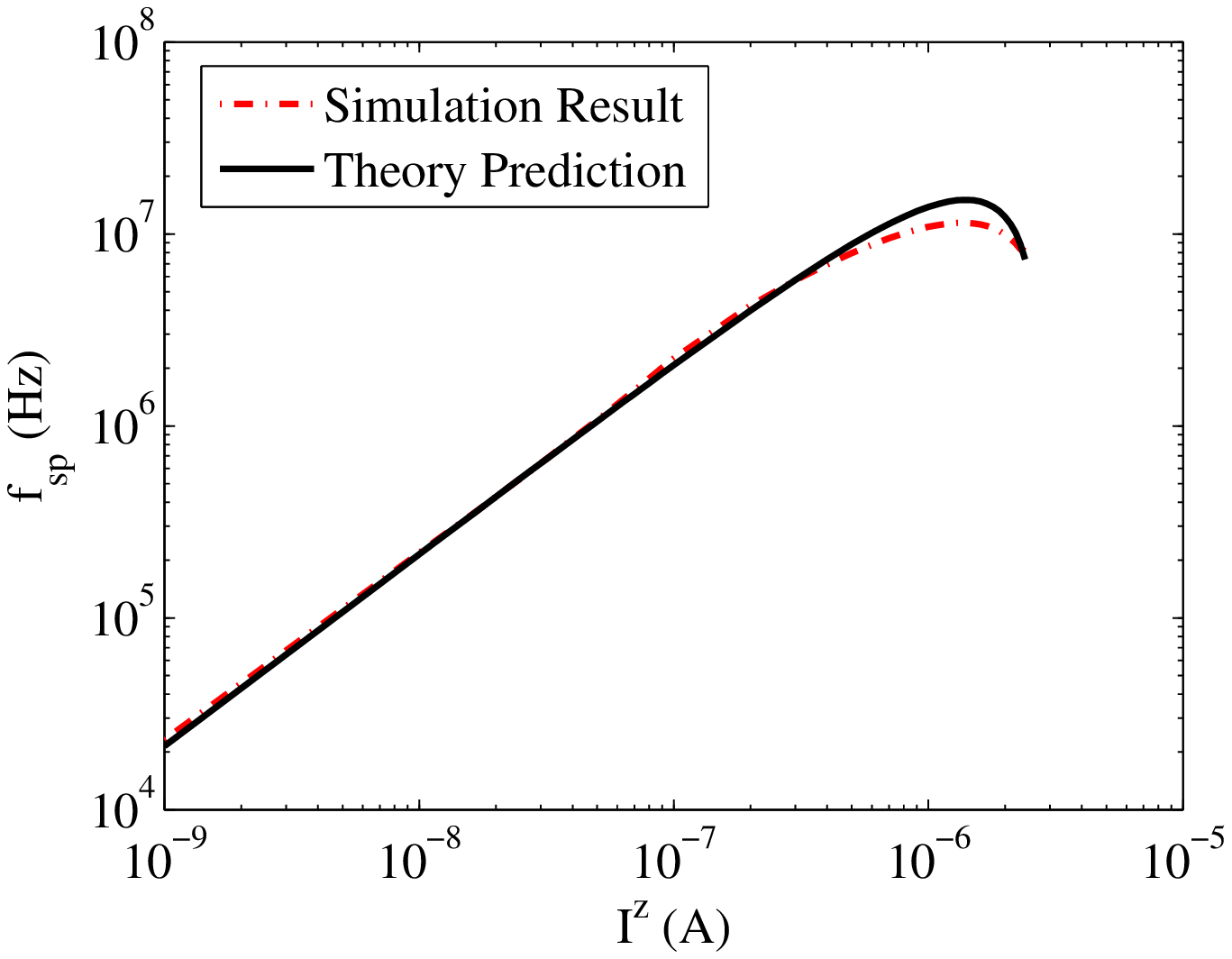} \\(a)\\
 	\includegraphics[width=0.35\textwidth]{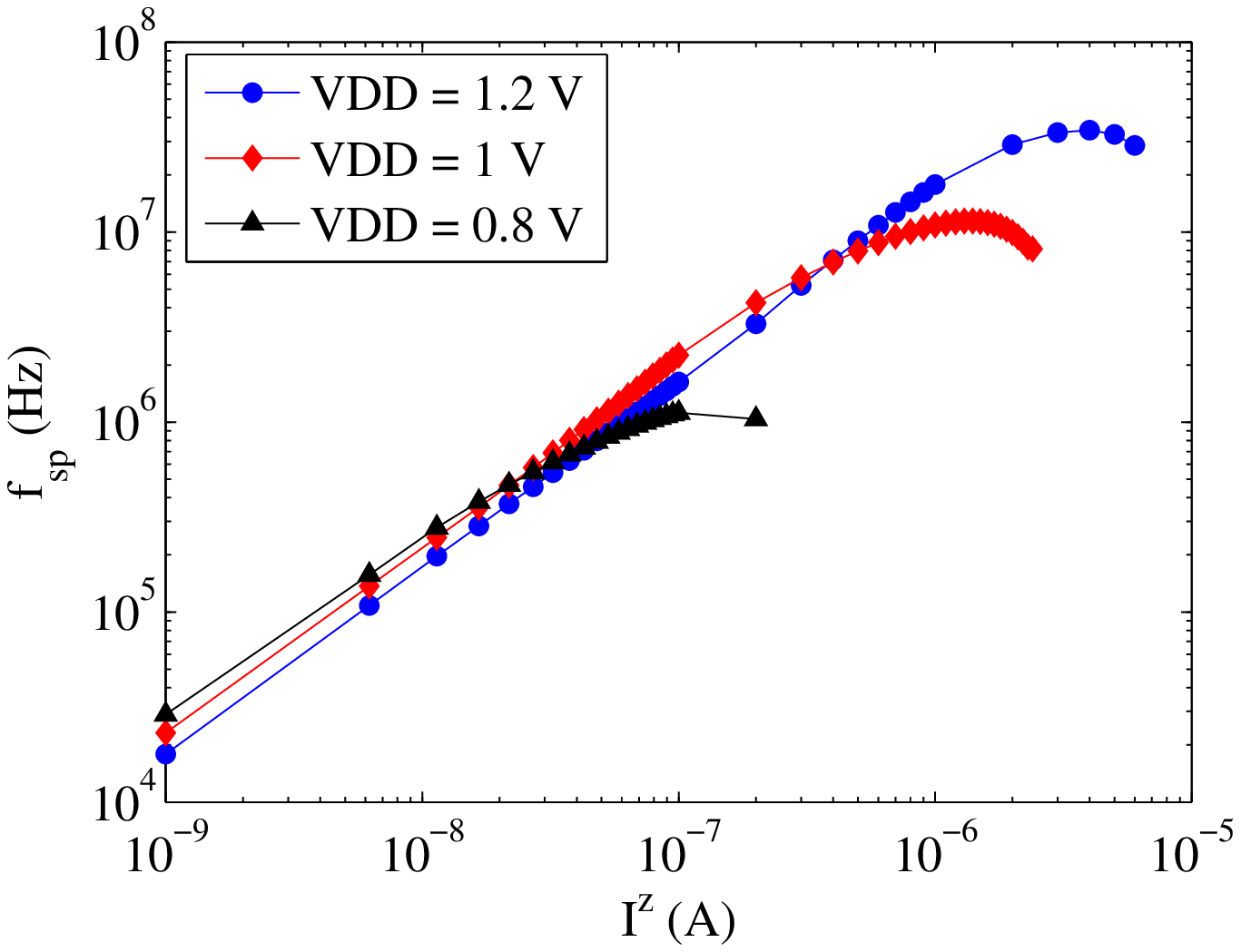} \\(b)
 	\caption{(a) Comparison of neuron spiking frequency between theory and simulation in SPICE show close match. (b) Simulated neuron spiking frequency with increasing input current for 3 different VDD. The curves saturate at higher maximum frequencies for higher VDD. Note the logarithmic scales for both plots.}\label{fig:fsp}
 \end{figure}

\subsection{Neuron}
\label{sec:neuron}
Figure \ref{fig:vmem}(a) details the circuit of the hidden layer neuron block. It is a current-controlled oscillator structure followed by an asynchronous counter. This is one of the simplest neuron circuits described in \cite{giacomo_neuron_review}. This circuit has the issue of large short-circuit current dissipation in the inverters. However, in our case we can avoid this problem by operating at very low power supply voltages ($\approx V_{TN}+V_{TP}$) making the short-circuit current negligible. The neuron is enabled when the control signal $NEU\_EN$ is high. The oscillation waveform at the nodes $V_{mem}$ and $V_{out}$ are illustrated in Fig. \ref{fig:vmem}(b). $V_{mem}$ is charged down by the input current $I^z-I_{lk}$ till it reaches the threshold voltage of the inverters. At that point both the inverters trip making the output switch to ground. Since the voltage change at the node of $V_{out}$ is VDD, the voltage change of $V_{mem}$ due to the feedback capacitor is given by:
\begin{equation}
\label{vmem}
\Delta V_{mem}=\frac{C_b}{C_a+C_b}VDD.
\end{equation}
Also, the reset transistor turns ON charging $V_{mem}$ up by the current $I_{rst}+I_{lk}-I^z$. The inverters trip again once $V_{mem}$ reaches the threshold and this process continues as long as $NEU\_EN$ is high. Both the capacitors $C_a$ and $C_b$ can be digitally reconfigured as shown in Fig. \ref{fig:vmem}(a). The values of the capacitors are: $C_{a1}=100$fF, $C_{a2}=200$fF, $C_{b1}=50$fF, $C_{b2}=100$fF.

We can derive an equation for the oscillation period $T_{sp}$. It is composed of two parts: the time $T_1$ for the input current $I^z$ to discharge the capacitor of node $V_{mem}$ and the time $T_2$ to reset the capacitor. Hence, $T_{sp}$ is given by:
\begin{equation}
\label{eq:tsp}
T_{sp}=T_1+T_2=C_bVDD\left(\frac{1}{I^z-I_{lk}}+\frac{1}{I_{rst}-I^z+I_{lk}}\right).
\end{equation}
Assuming $I_{lk}\approx 0$, the relationship between the neuron spiking frequency and the input current $I^z$ can be easily obtained as:
\begin{equation}
\label{eq:fsp}
f_{sp}=g\left(I^z\right)=\frac{I^z\left(I_{rst}-I^z\right)}{I_{rst}C_bVDD}.
\end{equation}
This quadratic relationship of equation (\ref{eq:fsp}) between current and frequency is plotted in Fig. \ref{subfig:transfer_func_a}. As we can see from Fig. \ref{subfig:transfer_func_a}, if $I^z << I_{rst}/2$, we have almost a linear relation given by:
\begin{equation}
\label{fsp_linear}
f_{sp}\approx\frac{I^z}{C_bVDD}=K_{neu}I^z,
\end{equation}
\begin{equation}
\label{Kneu}
K_{neu}=\frac{1}{C_bVDD}.
\end{equation}
where $K_{neu}=\frac{1}{C_bVDD}$ denotes a conversion gain from current to frequency. When $I^z = I_{rst}/2$, $f_{sp}$ will reach its maximum value $f_{max}$. After this point, the spiking frequency will keep falling down till it reaches zero for $I^z = I_{rst}$. Since the inflection point of the curve is reached at $I^z=I_{rst}/2$, we refer to this current value as $I_{flx}$. The chip has digital control bits making the capacitors configurable. As shown in Fig. \ref{fig:vmem}(a), an asynchronous counter counts the total number of spikes from the neuron  during a fixed period of time $T_{neu}$ (time duration for which $NEU\_EN$ is high) and generates the output $H$. A hard nonlinearity in the form of saturation can be implemented by stopping the counter whenever its count reaches a pre-defined limit $2^b$. $b$ in this case is the valid MSB of the counter output which is also configurable from $6$ to $14$. If only the linear region of the neuron spiking waveform is adopted (this is also the most energy efficient part as shown later), the final transfer function of the hidden layer neuron can be represented by:
\begin{equation}
\label{eq:H_function}
H = 
\begin{cases}
f_{sp}T_{neu}(\approx K_{neu}I^zT_{neu} if I^z<I_{flx}), & \mbox{if } H < 2^b \\
2^b. & \mbox{otherwise}
\end{cases}
\end{equation}
This saturating nonlinearity is shown in Fig. \ref{subfig:transfer_func_b}. This nonlinearity was preferred due to its ease of implementation and digital control. From Fig. \ref{subfig:transfer_func_b} we can also note the current at which the H saturates is denoted by $I^z_{sat}$. This value depends on both $T_{neu}$ and $b$. Also, $[0$ $I^z_{max}]$ is used to denote the range of input currents to the neuron.

  \begin{figure*} 
  	\begin{minipage}[b]{0.33\textwidth}
  		\centering
  		\includegraphics[width=1\textwidth]{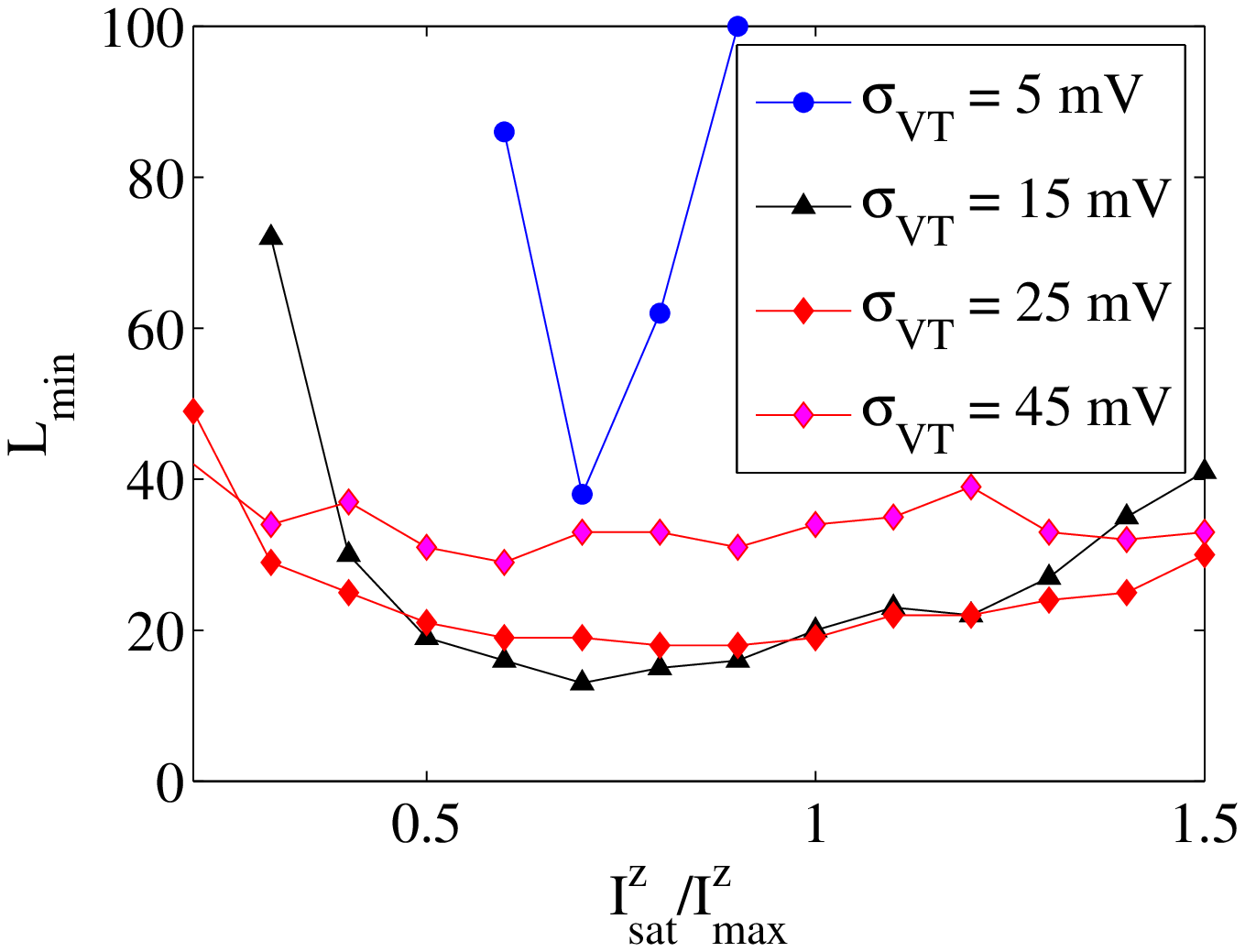} \\ (a)
  	\end{minipage}
  	\begin{minipage}[b]{0.33\textwidth}
  		\centering
  		\includegraphics[width=1\textwidth]{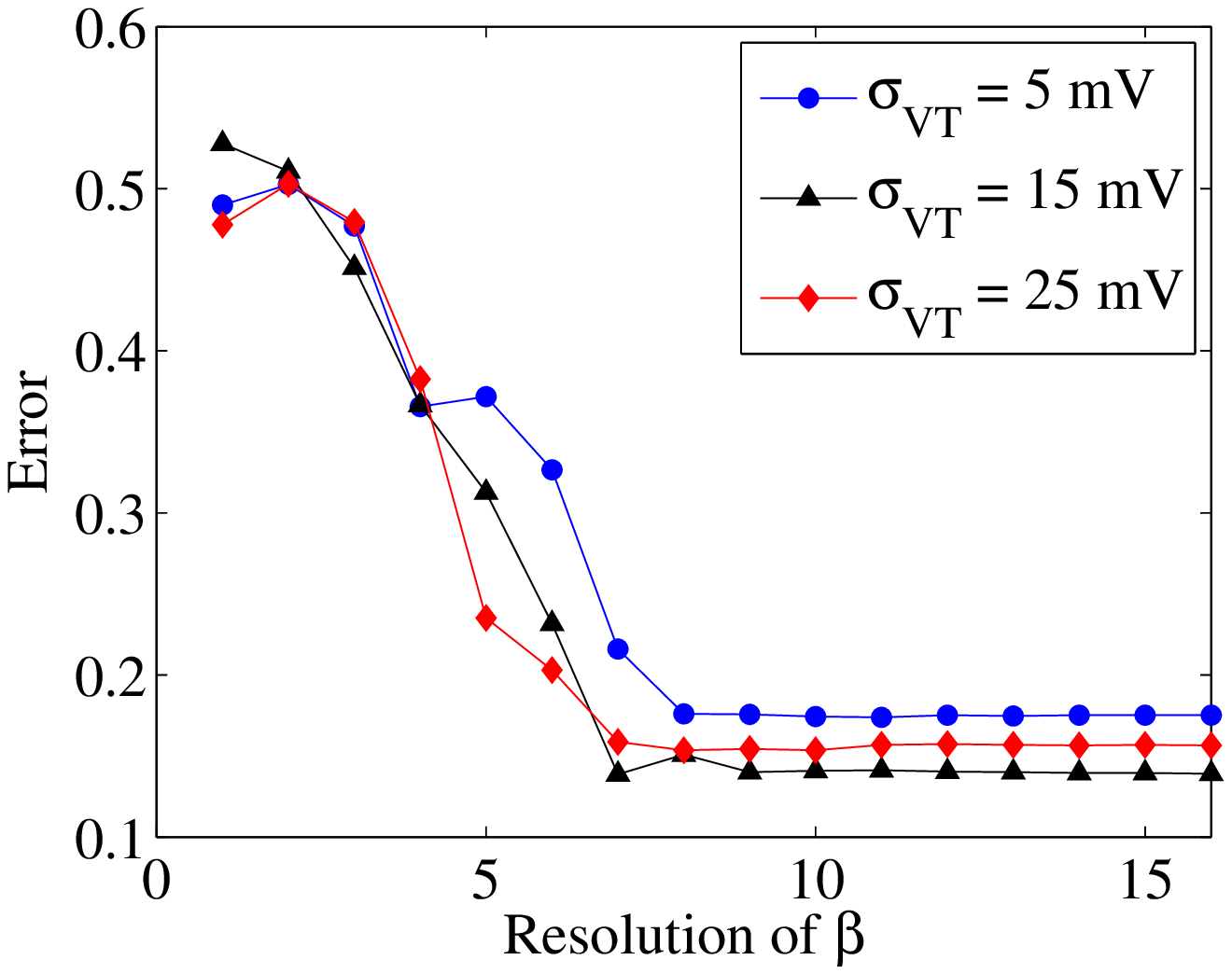} \\ (b)
  	\end{minipage}
  	\begin{minipage}[b]{0.33\textwidth}
  		\centering
  		\includegraphics[width=1\textwidth]{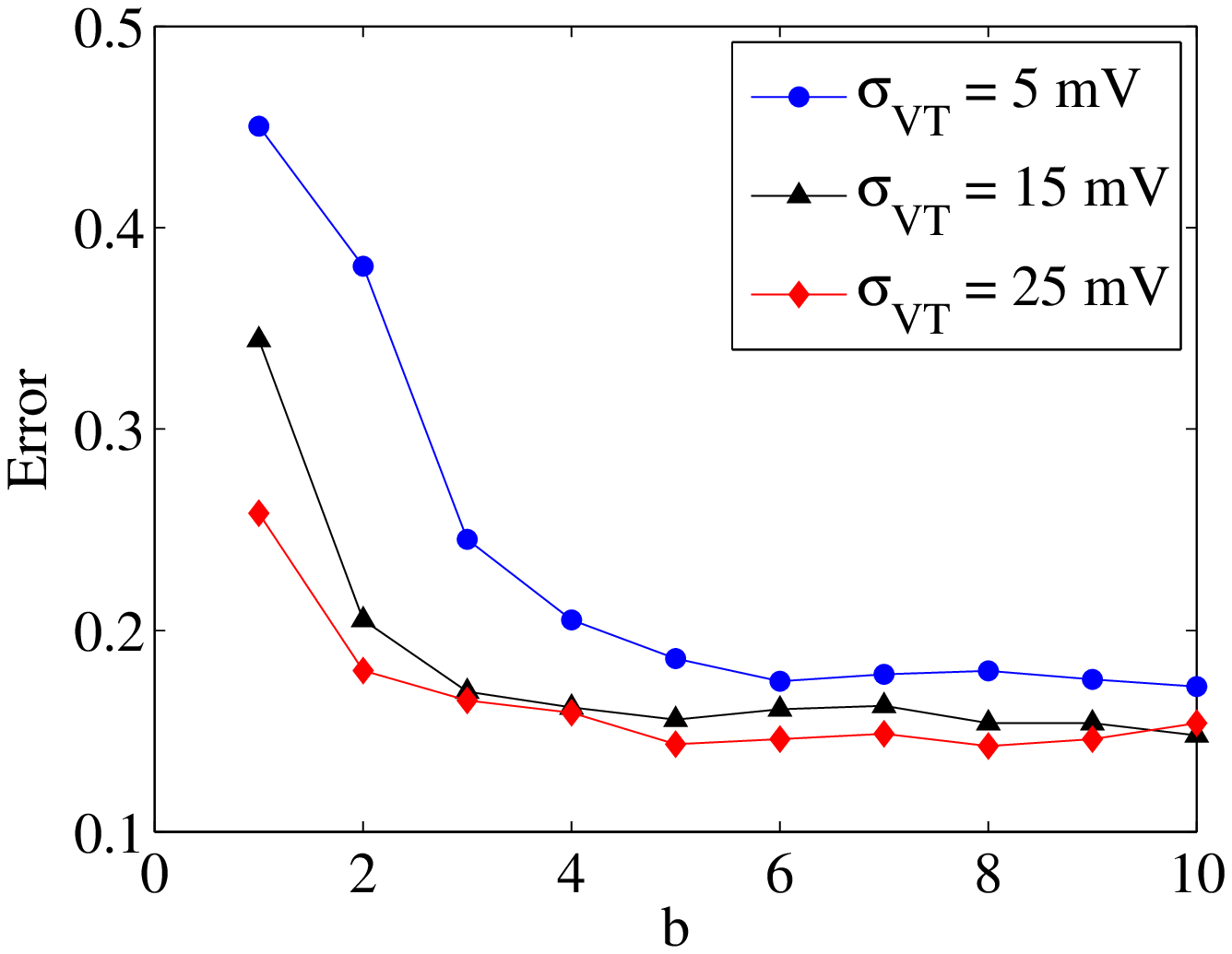} \\ (c)
  	\end{minipage}
  	\caption{\textbf{Design Space Exploration:} (a) Variations of $L_{min}$ with $I^z_{sat}/I^z_{max}$ show that the optimal value of this ratio is $\approx 0.75$. (b) Variations of classification accuracy with the resolution of output weight $\beta$ showing $10$ bits is sufficient for accurate classification. (c) Variations of classification accuracy with the number of bits of counter output $H$ demonstrating that $b\approx 6$ is enough for optimal performance. Each of the curves are averaged over $50$ trials.}
  	\label{fig:param_range}
  \end{figure*}
  
Figure \ref{fig:fsp}(a) plots SPICE simulation of the neuron spiking frequency with the variation of input current $I^z$ on a logarithmic scale and compares it with theoretical predictions based on equation \ref{eq:fsp}. For this simulation, $C_a$ and $C_b$ were set to be $300$fF and $50$fF respectively while VDD  was kept at $1$V. As expected, the spike frequency increases linearly for small values of $I^z$, reaches a maxima eventually and then starts reducing for further increase in $I^z$. Results from a similar simulation but for three different values of VDD ($0.8$, $1$ and $1.2$ V) are shown in Fig. \ref{fig:fsp}(b). Since $f_{sp}$ is inversely proportional to VDD, $f_{sp}$ is higher for small $I^z$ with a smaller VDD. However, when VDD is lower, $I_{rst}$ is smaller and hence $f_{sp}$ attains the peak value at smaller value of $I^z$, i.e. $I_{flx}$ reduces when VDD is reduced. On the other hand, for higher VDD, $f_{sp}$ saturates at a larger value $f_{max}$ and it is attained for larger value of $I_{flx}$.
 
\subsection{Current Mirror Array}
\label{sec:mirror}
The digital input $Data\underline{\ }in$ is mapped to a vector of input current $\textbf{I}_{in}$ which are copied to every neuron using a current mirror. These inputs can also be obtained from a sensor such as a photo diode. The capacitor $C=0.4$pF is kept to maintain a minimum SNR \cite{shantanu_mvm} at the expense of bandwidth. For low-power operation, we operate the current mirrors in sub-threshold regime. Minimum sized transistors are employed in these current mirrors to exploit VLSI mismatch which is necessary for the generation of random input weights $w_i$ and bias $b_i$ of ELM. For example, the contribution of input $i_{in,i}$ to the total input current of neuron $j$ is given by $i_{in,i}w_0e^{\Delta V_{T,ij}/U_T}$ where $U_T$ is the thermal voltage, $w_0$ is the nominal current mirror gain while $\Delta V_{T,ij}$ denotes the mismatch of the threshold voltage for the transistor copying the $i$-th input current to the $j$-th neuron. This last term is a random variable with a Gaussian distribution and hence the weights $\textbf{w}$ in equation ($\ref{eq:elm_eqn}$) above get mapped to random variables with a \emph{log-normal} distribution in our implementation. Since in our implementation $w_0=1$, we can write:
\begin{align}
\label{eq:weights}
w_{ij}=e^{\frac{\Delta V_{T,ij}}{U_T}}
\end{align}
Do note that the ELM algorithm only requires random numbers from any continuous distribution\cite{huang_elm}. Here ,we choose log-normal distribution due to the intrinsic physics of sub-threshold mosfets. If biased in above-threshold regime, the distribution of random numbers would be closer to gaussian.
 \begin{figure}[h]
 	\centering 	
 	\includegraphics[width=0.25\textwidth]{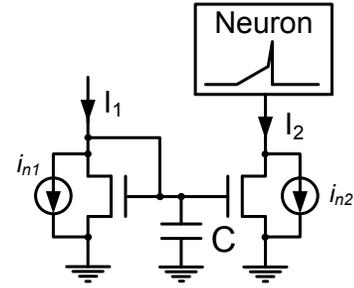}
 	\caption{Simplified circuit diagram of one current mirror for noise analysis.}\label{fig:noise_cm}
 \end{figure}

\subsection{Parameter Choice}
\label{sec:parameter}
To determine the performance of the network, we chose two representative tasks of regression ($d=1$) and classification ($d=14$). For the regression task, the network was given a set of noisy samples and had to approximate the underlying function. For classification, six different data sets with widely varying dimensions and training set sizes were chosen from the UCI machine learning repository\cite{uci}. Here, we show results for only the `brightdata' case as a representative but the conclusions drawn are valid across the other data sets. It is a two class problem that includes $1000$ training data and $1462$ testing data. The reasons for choosing these tasks were that the performance of the software implementation for these tasks are reported in publications as a typical benchmark\cite{huang_elm_kernel}.

For the following simulations done in MATLAB, we considered the mismatch in current mirror weights as the dominant factor. It was assumed to be log-normally distributed with a standard deviation of $V_T$, $\sigma_{V_T}$ ranging from $5$ to $45$ mV (as a reference, $\sigma_{V_T}$ in our fabricated chip is $\approx 16$ mV). Equation (\ref{eq:H_function}) was used to simulate the neuronal characteristic and the other parameters were kept at fixed nominal values of $K_{neu}=26$KHz/nA and $T_{neu}=56 \mu$sec. In real applications, variations exist for other parameters in the neuron transfer function as well. However, simulation results show that mismatch in these	 do not affect the qualitative nature of the results we present here. 

  \begin{figure*} 
  	\begin{minipage}[b]{0.33\textwidth}
  		\centering
  		\includegraphics[width=1\textwidth]{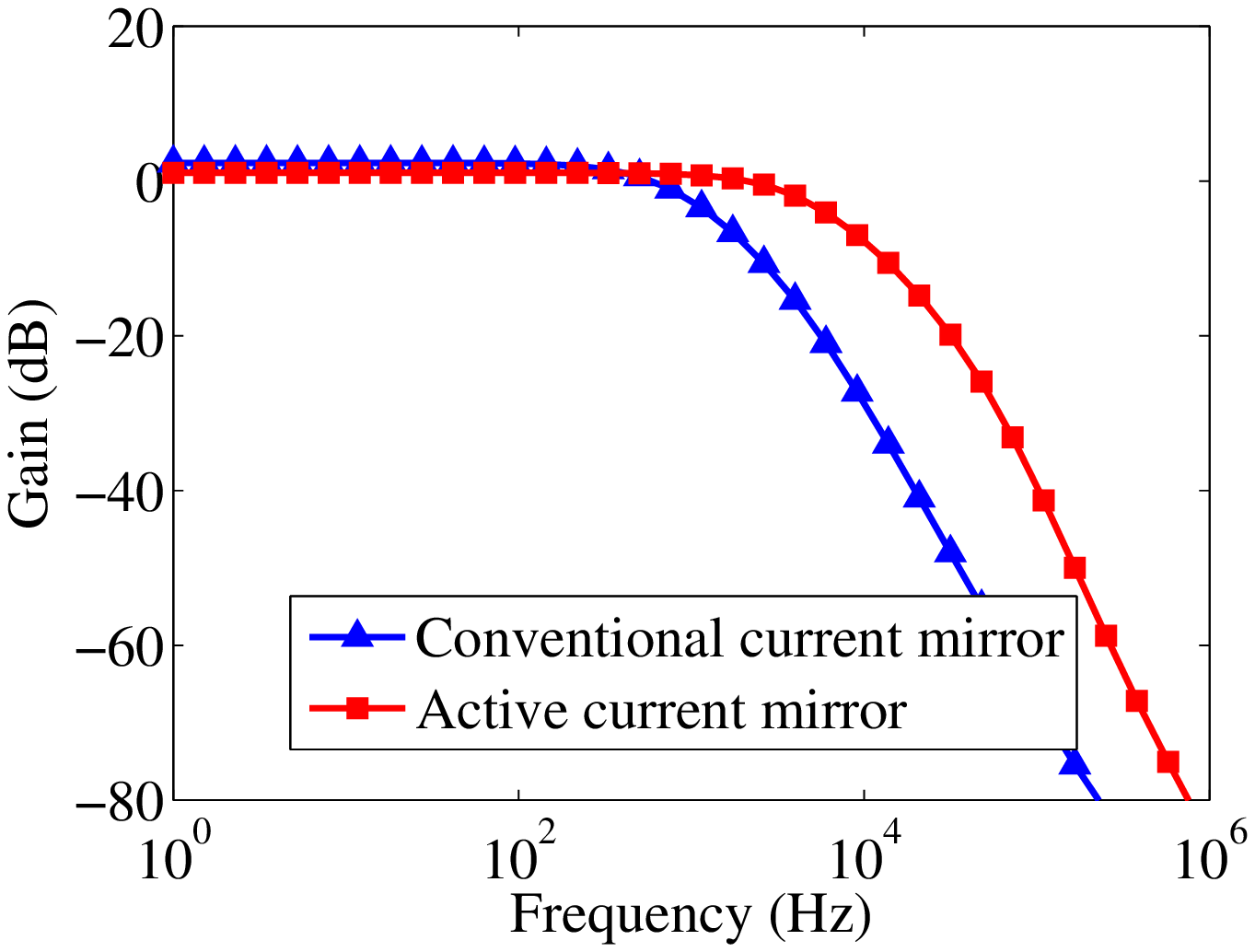} \\ (a)
  	\end{minipage}
  	\begin{minipage}[b]{0.33\textwidth}
  		\centering
  		\includegraphics[width=1\textwidth]{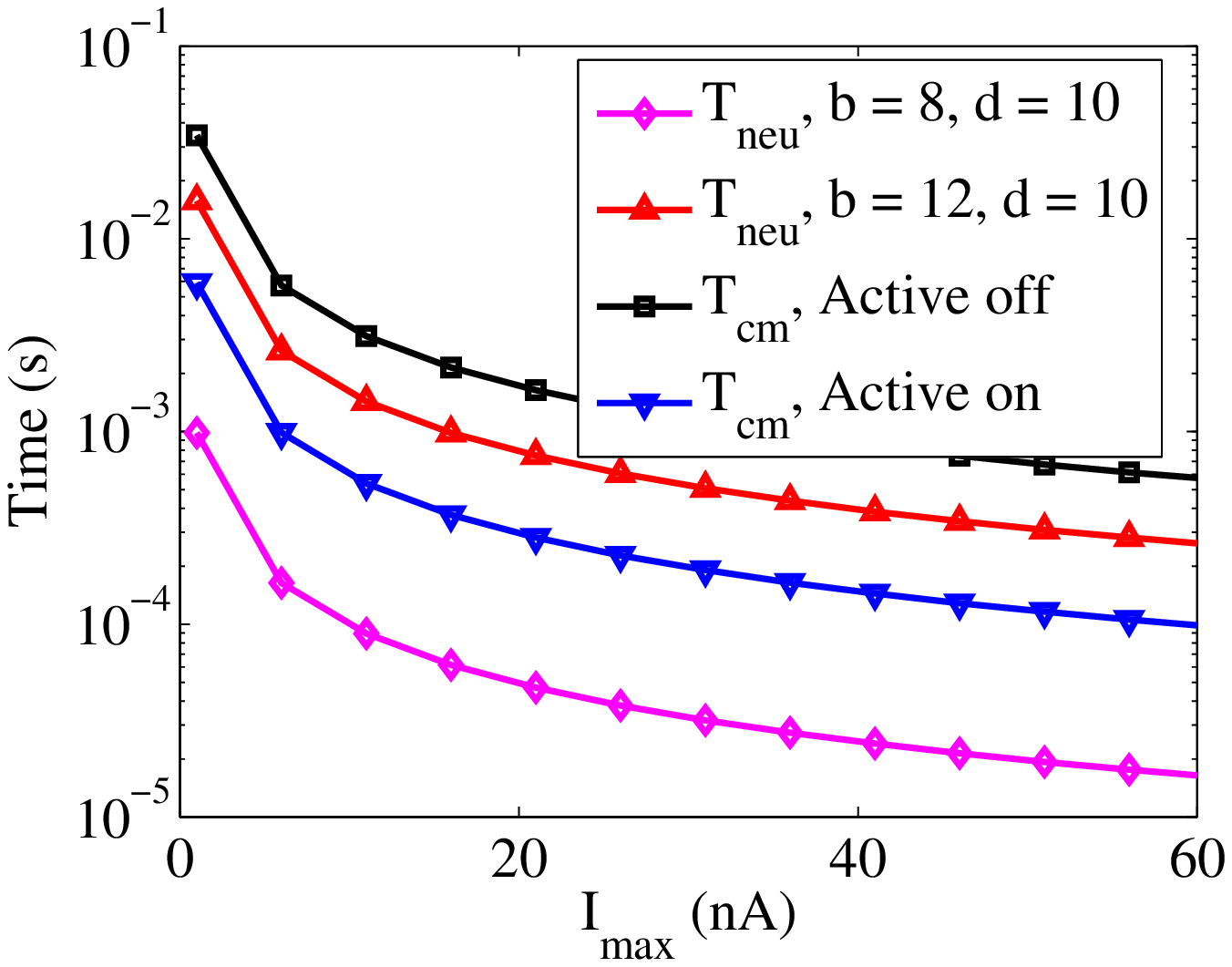} \\ (b)
  	\end{minipage}
  	\begin{minipage}[b]{0.33\textwidth}
  		\centering
  		\includegraphics[width=1\textwidth]{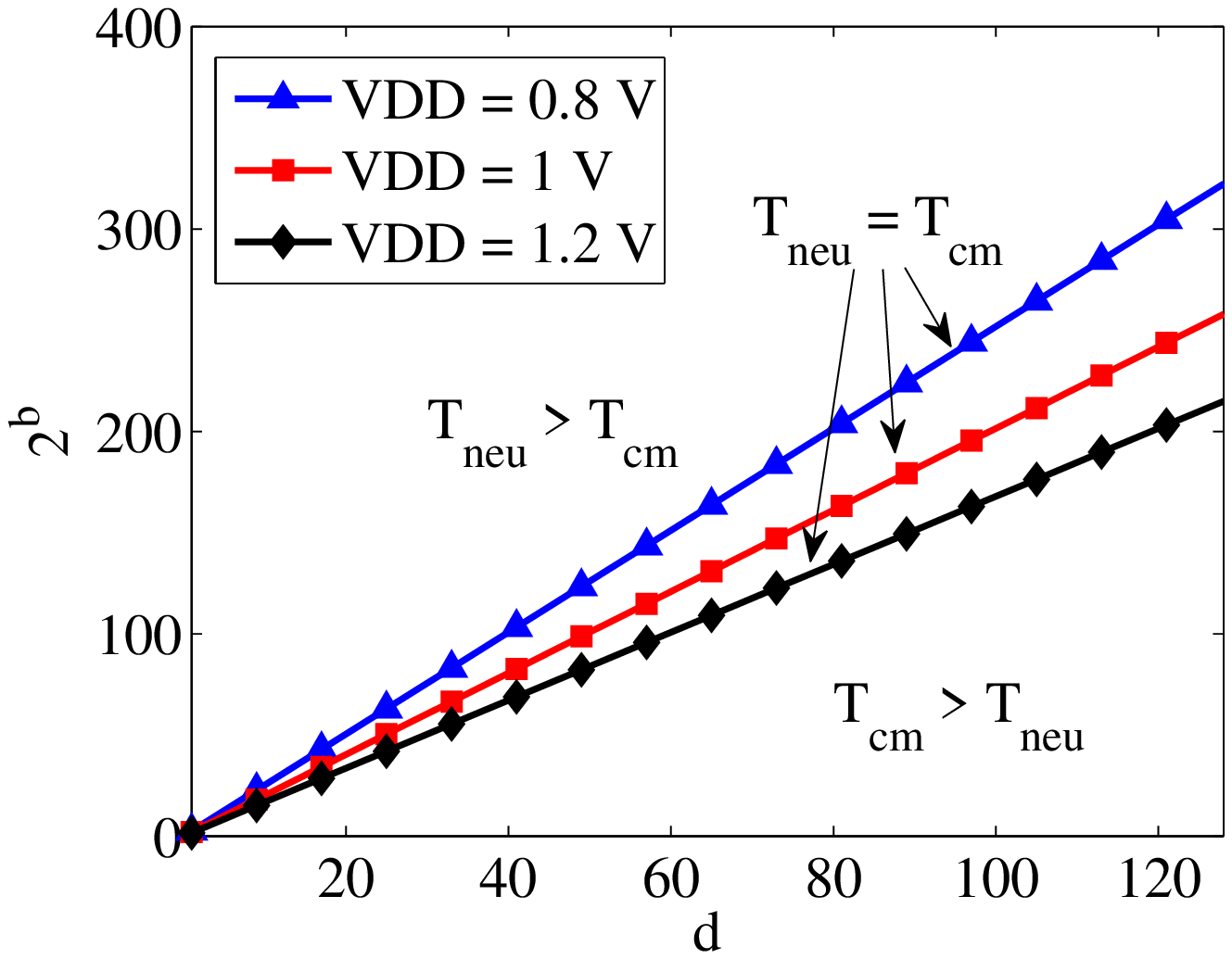} \\ (c)
  	\end{minipage}
  	\caption{\textbf{Trade-offs in speed:} (a) Using active current mirror for small input currents can boost the bandwidth by $5.84X$. (b) Variation of neuron counting time ($T_{neu}$) and current mirror settling time ($T_{cm}$) reduce as maximum input current per dimension ($I_{max}$) is increased. Further, $T_{neu}$ increases exponentially with increase in $b$. (c) Contours where $T_{cm}$ is equal to $T_{neu}$ in the space of counter dynamic range $2^b$ and input dimension $d$. For increasing $d$, the total current input to a neuron $I^z$ keeps on increasing thus increasing oscillation frequency. Hence, it can support higher dynamic range $2^b$ in the same time $T_{neu}$.}
  	\label{fig:active_current_mirror}
  \end{figure*}
  
\subsubsection{Input Mapping}
For efficient use of the hardware, we need to determine how to map the compact set $X=[-1$ $1]$ to input currents. First, it can be only mapped to a set in $R^+$ since we have unidirectional current mirrors. Assume the maximum input current for one dimension is $I_{max}$, i.e., the set is $[0$ $I_{max}]$. Therefore the maximum current going to the neuron $I^z_{max} = d\times I_{max}$. From Fig. \ref{subfig:transfer_func_b}, we need to find out the relationship between $I^z_{max}$ and $I^z_{sat}$. Though theoretically any positive set will work, it might need an unreasonably large number of neurons to get a satisfactory performance. To illustrate this point intuitively, consider a case where $I^z_{max}<<I^z_{sat}$. Then the transfer function of the neuron is a linear function without any high order components. Also, if $I^z_{max}>>I^z_{sat}$, the outputs of most neurons will be saturated to $2^b$, and will not encode the variations of the input. Both of these cases will require a large number of hidden layer neurons so that `by chance' a large enough pool of neurons are obtained which encode the changes in input. Hence, there should be a range for the ratio between $I^z_{max}$ and $I^z_{sat}$, such that  we can achieve a good performance with a small number of hidden layer neurons.

To find this desired range, we first fix a value of $I^z_{sat}/I^z_{max}$ and evaluate the performance of the network on both tasks with different number $L$ of hidden layer neurons. The regression error reduces initially with larger $L$ but saturates after the $L$ increases beyond a critical value $L_{min}$. 
To quantify the dependence of performance on the ratio of $I^z_{sat}/I^z_{max}$ , we now plot in Fig. \ref{fig:param_range}(a) the dependence of $L_{min}$ on the ratio of $I^z_{sat}/I^z_{max}$ , with lower values of $L_{min}$ being preferable. We have chosen error of $0.08$ as the saturation level in this case. From this figure, the ratio of $I^z_{sat}/I^z_{max} \approx 0.75$ is the best trade off point between number of hidden neurons and input dynamic range for all values of $\sigma_{V_T}$. For small values of $\sigma_{V_T}$, the performance degrades rapidly on both sides of the optimal value. However, as $\sigma_{V_T}$ increases, the performance degradation is much less implying the choice of $I^z_{max}$ is less critical in highly scaled VLSI. 

However, it can also be noted that the performance is best (least $L_{min}$) for $\sigma_{V_T}$ in the range of $15-25$mV. This has been found to be true for a wide range of classification problems as well. Hence, for deeply scaled CMOS processes with larger $\sigma_{V_T}$, minimum sized transistors cannot be used. In those cases, the transistor size has to be increased (following Pelgrom's model\cite{kinget-mismatch}) to reduce $\sigma_{V_T}$ within the desired range. However, the required area will still reduce compared to an older process with larger transistors since the coefficient $A_{V_T}$ is reducing as transistor scaling continues\cite{kinget-mismatch}.

\subsubsection{Resolution of Output Weight}
As mentioned earlier, the digital circuits will use pre-calculated output weights, $\beta$ from a memory and accumulate it based on neuronal spiking patterns. In order to implement this, we need to know how many bits are needed to represent $\beta$. Less number of bits will degrade performance of the classifier while more will waste hardware resources and power. We use the classification example here with $L=128$. Figure \ref{fig:param_range}(b) shows the change of error with increasing number of bits indicating $10$ bits resolution is enough for good accuracy.

\subsubsection{Counter resolution}
Besides the resolution of $\beta$, we also analyzed the dependence of performance on the output counter resolution $b$ in equation (\ref{eq:H_function}). Since we estimate the spiking frequency by using a counter to count the number of spikes in a fixed time window $T_{neu}$, a small value of $b$ will introduce large quantization errors in the estimate of frequency. This implies that the neurons have to produce more spikes in the counting window, which would on the other hand induce more power dissipation. To find a good trade-off for $b$, we fixed $I^z_{sat}/I^z_{max} \approx 0.75$, $L=128$ and resolution of $\beta$ to $10$ bits. Figure \ref{fig:param_range}(c) shows the simulation result for the classification error with $b$ increasing from $1$ to $10$. $b \approx 6$ is found to be sufficient for classification.

\section{Noise, Speed and Energy dissipation}
\label{sec:energy}
\subsection{Noise}
Noise is an important specification to be considered in circuit design. In this section, we present the operational limits set on this architecture due to noise based constraints. Since the transistors are operating in sub-threshold region, the contribution of $1/f$ noise is negligible compared to the thermal noise\cite{shantanu_mvm}. For the current mirror circuit as shown in Fig. \ref{fig:noise_cm}, we can easily get the input referred thermal noise spectral density as:
\begin{equation}
\overline{i_{in}^2}=\overline{i_{n1}^2}+\overline{i_{n2}^2}\cdot\frac{g_{m1}^2}{g_{m2}^2},
\end{equation}
where $g_{m1}$ and $g_{m2}$ are transconductance of input and output transistors respectively, $\overline{i_{n1}^2}$ and $\overline{i_{n2}^2}$ are corresponding transistor channel noise. Since the transistors are working in the sub-threshold region, the transconductance is in proportion to its drain current. Applying the noise model of drain current of sub-threshold transistors to be $\overline{i^2}=2qI{\Delta}f$\cite{sarpeshkar} where $q$ denotes the electronics charge, we can rewrite the above equation as:
\begin{equation}
\overline{i_{in}^2}=2qI_1{\Delta}f+2q{\Delta}f\cdot\frac{I_1^2}{I_2}
\end{equation}
For this single pole system, the noise equivalent bandwidth $\Delta f=\frac{{\kappa}I_1}{4CU_T}$ where $\kappa$ denotes the inverse of the sub-threshold slope\cite{sarpeshkar}. Assuming $I_2/I_1=w_0$, and substituting the bandwidth equation above, we get:
\begin{equation}
\overline{i_{in}^2}=\frac{q\kappa I_1^2}{2CU_T}\left(1+\frac{1}{w_0}\right).
\end{equation}
Finally, the signal to noise ratio (SNR) can be expressed in the following equation:
\begin{equation}
\label{eq:snr}
SNR=\frac{I_1^2}{\overline{i_{in}^2}}=\frac{2CU_Tw_0}{q\kappa (w_0+1)}.
\end{equation}

Thus, from the equation (\ref{eq:snr}), we can see the SNR can be controlled by changing $C$. This reflects a direct trade-off with bandwidth which is inversely proportional to $C$. If an 8 bits SNR is needed in the system, and $w_0=1$, it is sufficient to add $C=0.4$pF capacitance in the current mirror for each input channel. Note that only one such capacitor is needed for every row.
 \begin{figure}[h]
 	\centering 	
 	\includegraphics[width=0.4\textwidth]{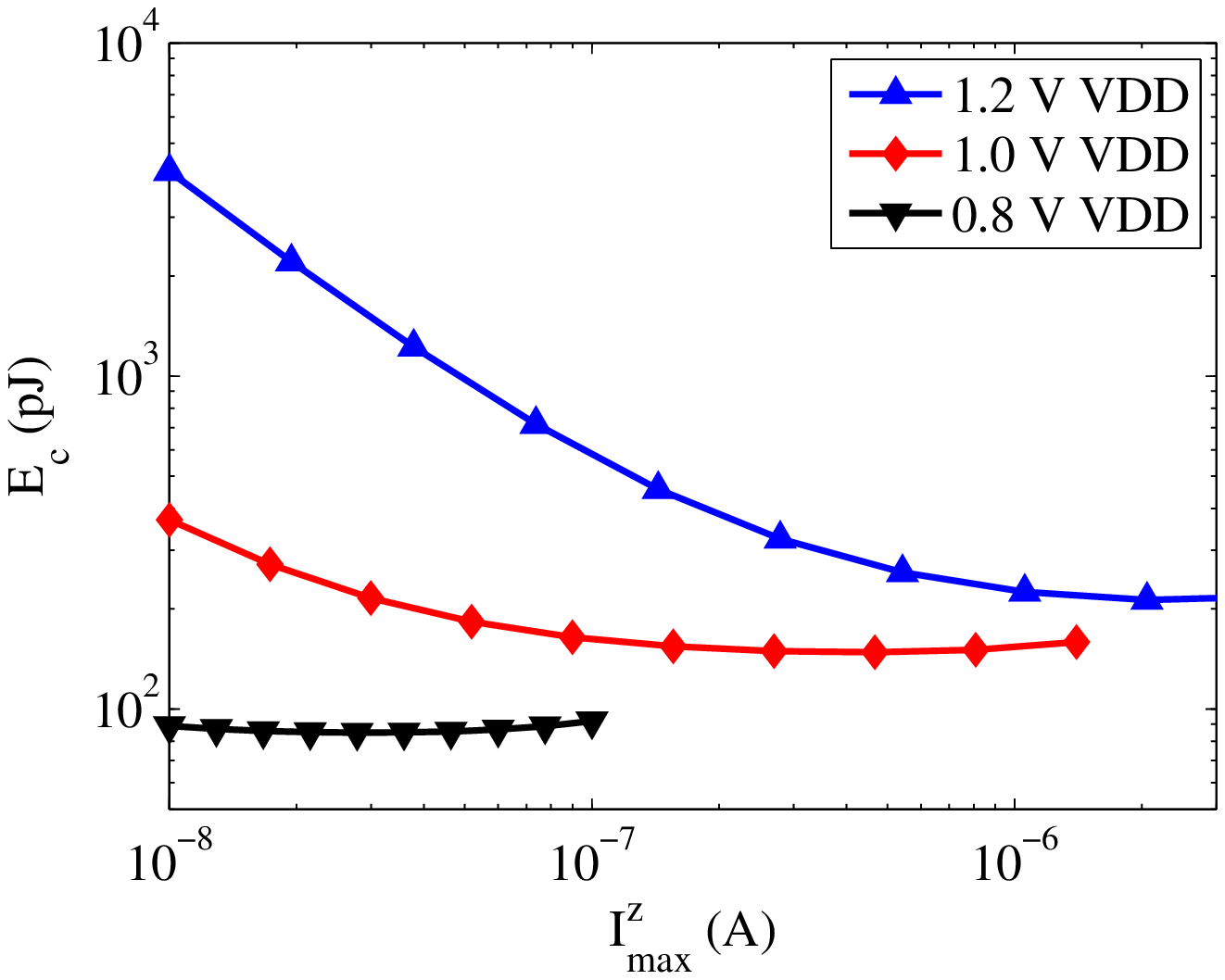} \\(a)\\
 	\includegraphics[width=0.4\textwidth]{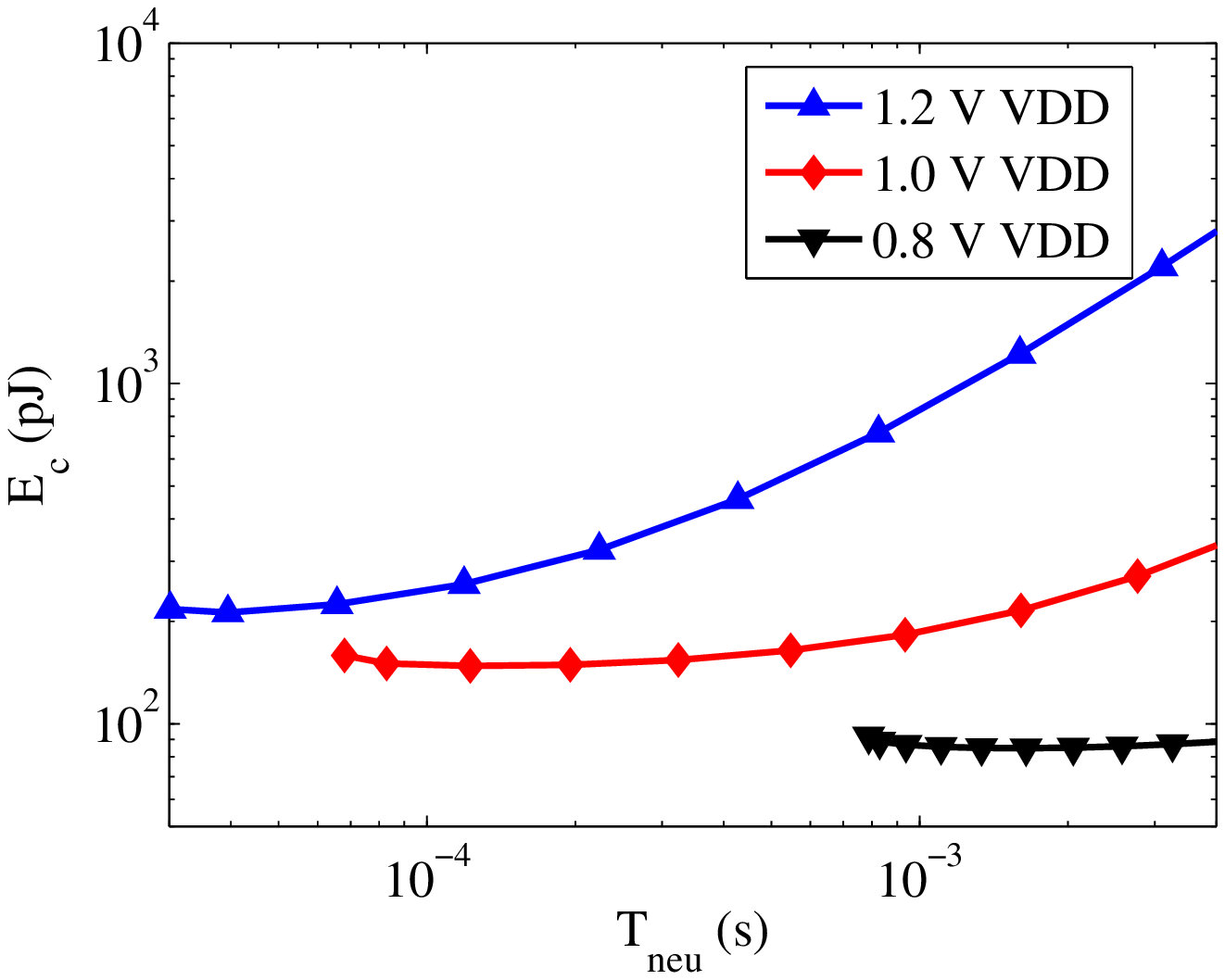} \\(b)
 	\caption{(a) Variation of energy per classification operation ($E_c$) with varying maximum value of input current $I^z_{max}$ for three different settings of VDD. (b) The same plot as in (a) but replacing $I^z_{max}$ with its corresponding $T_{neu}$ from equation (\ref{eq:tneu}).}\label{fig:Ec_Imax}
 \end{figure}
 
\subsection{Speed}
The conversion time for one classification operation $T_c$ comprises two parts: $T_{cm}$ and $T_{neu}$ where $T_{neu}$ is the neuron operation time and $T_{cm}$ is the current mirror settling time. If one of them is much larger than the other, we can approximate $T_c\approx max\left(T_{cm}, T_{neu}\right)$. We consider $T_{cm}$ to be $4$ times of the inverse of the bandwidth (BW), i.e. $T_{cm}=\frac{4}{BW}=\frac{4CU_T}{\kappa I_{in}}$ where $\kappa=0.7$, $U_T=0.025$V at room temperature and $C=0.4$pF as derived earlier. 
 If the average input current is $I_{max}/2$, the average current mirror settling time is
\begin{equation}
\label{eq:tcm}
T_{cm,avg}=\frac{8CU_t}{\kappa I_{max}}.
\end{equation}
As discussed earlier in Section \ref{sec:igc}, an active current mirror is utilized to boost the bandwidth for small current values. SPICE simulation result for this effect shown in Fig. \ref{fig:active_current_mirror}(a) demonstrates a bandwidth increase by around $5.84X$. We can find the range of $T_{cm}$ by considering maximum and minimum input currents:
\begin{align}
\label{eq:tcm_range}
T_{cm,max}&=\frac{4CU_t}{5.84\kappa I_{max}/2^{b_{in}}}\notag\\
T_{cm,min}&=\frac{4CU_t}{\kappa I_{max}}
\end{align}
where $b_{in}=10$ is the number of bits of $Data\_in$ and the factor of $5.84$ is due to the active current mirror. Figure \ref{fig:active_current_mirror}(b) shows the decrease of $T_{cm}$ with increasing $I_{max}$ for the conventional and active current mirror cases. 

To find the value of $T_{neu}$, we can see from Fig. \ref{fig:transfer_func}(b) that we want $H=2^b$ for $I^z=I^z_{sat}$. Combining this observation with equation (\ref{eq:H_function}), we can derive the following:
\begin{equation}
\label{eq:tneu}
T_{neu}=\frac{2^b}{K_{neu}I^z_{sat}}=\frac{2^b}{0.75K_{neu}I^z_{max}}=\frac{2^b}{0.75K_{neu}dI_{max}}.
\end{equation}
where we use $I^z_{sat}/I^z_{max} = 0.75$ (shown earlier in Section \ref{sec:parameter}) and $I^z_{max} = d \times I_{max}$. Now, we can compare $T_{cm}$ and $T_{neu}$ to see the dominant term as a function of parameters $b$ and $d$. 
Figure \ref{fig:active_current_mirror}(b) shows a comparison between $T_{cm}=0.5(T_{cm,max}+T_{cm,min})$ and $T_{neu}$ for $b=8$ and $b=12$. Increasing $I_{max}$ reduces the time required for both the neuron and current mirror. $T_{cm}$ for the conventional current mirror is always the dominant factor. However, with the active current mirror on $T_{neu}$ may be larger than $T_{cm}$ for large values of $b$. These plots are done for $d=10$; increasing $d$ will have an effect of reducing $T_{neu}$ since $I^Z_{max}=d\times I_{max}$ increases. Hence, to show the trade-offs between $T_{cm}$ and $T_{neu}$ as a function of $b$ and $d$, we plot contours in the space of counter dynamic range $2^b$ and input dimension $d$ where $T_{cm}=T_{neu}$. To do this, we equate (\ref{eq:tcm}) and (\ref{eq:tneu}) to get:
\begin{align}
\label{eq:contour}
\frac{8CU_t}{\kappa I^z_{max}/d} &=\frac{2^b}{K_{neu}I^z_{sat}}\notag\\
\implies 2^b &=\frac{6dCU_tK_{neu}}{\kappa}
\end{align}
where $I^z_{sat}/I^z_{max} = 0.75$ is used.
The straight line contours defined by equation \ref{eq:contour} are plotted in Fig. \ref{fig:active_current_mirror}(c) for three different $K_{neu}$  values corresponding to VDD$=0.8$, $1$ and $1.2$V. For parameter choices on these contour lines, $T_c = T_{cm}+T_{neu}= 2T_{cm} = 2T_{neu}$. If the relation between $2^b$ and $d$ sets the operation regime above any of the contour lines, $T_{neu} > T_{cm}$ while the opposite condition is true if operation regime is below the contour lines. It can be seen that for $b\approx 8-10$ bits and a nominal value of VDD=$1$V, $T_{neu}$ dominates $T_{cm}$ for the maximum dimension of $128$ supported by our chip.

\subsection{Energy}
\label{sec:energy- model}
The total power dissipated by the system ($P_t$) can be split into two parts: power from analog ($P_{avdd}$) and digital ($P_{vdd}$) supplies. 
The first term ($P_{avdd}$) is mainly dissipated by the voltage reference circuitry, biasing block and the IGCs. Ideally, this should be a function of input dimension. However, in the current design only unused active mirrors are turned off while the current DAC is always ON--this will be rectified in future designs. The second term ($P_{vdd}$) comprises the power dissipated by the neuron, asynchronous counter and other digital blocks including decoder and scanner. Of these terms, the power dissipated by the neuron includes the synaptic currents as the input and the counter at output and varies with different parameters such as biasing current. It is the major energy consumer in the chip when the number of hidden neurons, $L$ is large. Hence, it is important to understand its dependence on different parameters. Thus, we can write $P_{vdd}$ as:
\begin{equation}
\label{eq:pvdd}
P_{vdd}=P_{neu}+P_{dig}\approx P_{neu}=Lf_{sp}E_{sp},
\end{equation}
where $E_{sp}$ is the energy dissipation per spike for the neuron.   $E_{sp}$ can be modelled as:
\begin{equation}
\label{eq:esp}
E_{sp}=\alpha_1VDD^2+\frac{\alpha_2I_{sc}VDD}{f_{sp}}+\frac{C_bI^zVDD^2}{I_{rst}-I^z+I_{lk}},
\end{equation}
where $I_{sc}$ is the short-circuit current in the inverter that depends on the value of VDD and is negligible for small values of VDD. Here, the first term denotes the switching power dissipated in the neuron circuit, second term denotes short circuit power loss in the inverters and the third term denotes the short-circuit power dissipated on the node $V_{mem}$ in Fig. \ref{fig:vmem}(a).
If $I^z << I_{rst}$ and $I_{lk}\approx 0$, equations (\ref{eq:pvdd}) and (\ref{eq:esp}) can be combined to give: 
\begin{equation}
\label{eq:psp}
P_{vdd}\approx P_{neu}\approx L\left(\alpha_1VDD^2f_{sp}+\alpha_2I_{sc}VDD\right).
\end{equation}
From simulation, when VDD is $1$V, $\alpha_1\approx 0.2 pF$ and $\alpha_2I_{sc}\approx 0.03 \mu A$.

Using equation (\ref{eq:esp}), we will now proceed to estimate average energy per conversion operation ($E_c$) for one neuron where an input current $I^z\in [0$ $I^z_{max}]$ is converted to a digital count. Assuming that $I^z$ is distributed uniformly in the range of $0$ to $I^z_{max}$, i.e. $P(I^z)=\frac{1}{I^z_{max}}$, $E_c$ can be estimated as:
\begin{align}
\begin{split}
\label{eq:energy_conv}
E_c & =\int_0^{I^z_{max}}E_{sp}\left(I^z\right)H\left(I^z\right)P\left(I^z\right)dI^z\\
& =\frac{1}{I^z_{max}}\int_0^{I^z_{max}}E_{sp}\left(I^z\right)H\left(I^z\right)dI^z, \\
\end{split}
\end{align}
where $H(I^z)$ is the number of spikes generated in $T_{neu}$ as defined in equation (\ref{eq:H_function}). Note that here we write $E_{sp}(I^z)$ and $H(I^z)$ to make the dependence of equations (\ref{eq:esp}) and (\ref{eq:H_function}) on $I^z$ explicit. Using the expression for $T_{neu}$ in equation (\ref{eq:tneu}), equation (\ref{eq:energy_conv}) can be simplified further to get:
\begin{equation}
\label{eq:energy_conv_simp}
E_c=\frac{2^b}{0.75K_{neu}{I^z_{max}}^2}\int_0^{I^z_{max}}E_{sp}\left(I^z\right)f_{sp}\left(I^z\right)dI^z.
\end{equation}
From equation (\ref{eq:energy_conv_simp}), we can see that $E_c$ depends on $I^z_{max}$. The choice of $I^z_{max}$ is guided by the design constraints. Typically, we have to either meet a minimum specified speed of operation or minimize energy of operation without any constraint on speed. To better explain the trade-offs, we can plot $E_c$ while varying $I^z_{max}$ with $b = 10$ as illustrated in Fig. \ref{fig:Ec_Imax}(a) for three values of VDD. The same figure is re-plotted in Fig. \ref{fig:Ec_Imax}(b) but with the corresponding value of $T_{neu}$ instead of $I^z$. Firstly, note that the plots for smaller VDD span a smaller range of current since $I_{rst}$ is correspondingly smaller (similar to Fig. \ref{fig:fsp}). For each VDD, the lowest conversion energy is attained when $I^z_{max}$ is close to $I_{flx}=I_{rst}/2$. Intuitively, this happens because $f_{sp}$ is higher which leads to lower $T_{neu}$ and correspondingly lower energy. Thus it is beneficial to operate for a short time at a higher spiking frequency than over a longer time with a small frequency. The optimum current $I^z$ is less than $I_{flx}$ since at $I^z=I_{flx}$, the short-circuit power dissipation (third term in equation (\ref{eq:esp})) increases significantly. From Fig. \ref{fig:Ec_Imax}, we can see that lowest energy per conversion is attainable for lowest VDD as expected since the short circuit current reduces drastically at lower VDD. However from Fig. \ref{fig:Ec_Imax}(b), we can see that the trade-off for keeping a low VDD is large conversion time. Hence, if conversion time is a critical specification, we have to choose the minimum VDD that meets this specification. As can be seen from Fig. \ref{fig:Ec_Imax}(b), higher VDD allows for lower $T_{neu}$. 

\begin{figure}[t]
	\centerline{
		\includegraphics[width=0.35\textwidth]{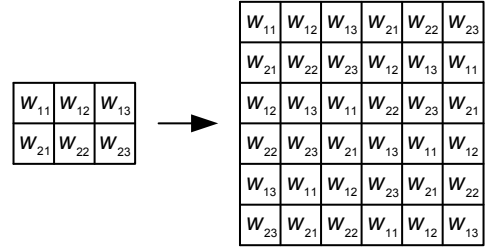} }
	\caption{The extension from a $2\times 3$ random projection matrix to $6\times 6$ by weight reuse technique .}\label{fig:weight_reuse}
\end{figure}

\begin{figure}[t]
	\centering
	\begin{minipage}{0.45\textwidth}
		\centering
		\subfigure[]{
			\label{subfig:schematic_hidden_layer_extension}
			\includegraphics[width=0.55\textwidth]{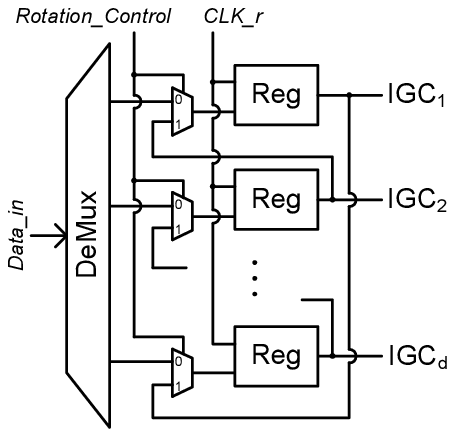}}
		\centering
		\subfigure[]{
			\label{subfig:timing_hidden_extension}
			\includegraphics[width=0.95\textwidth]{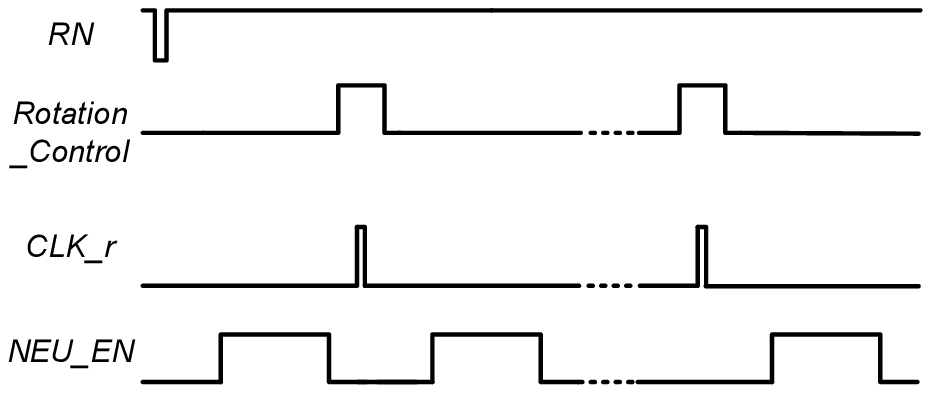}}
		\centering
		\caption{(a) Schematic of peripheral circuit for hidden layer extension by shifting the input data stored in the registers and (b) its timing diagram.}\label{fig:hidden_expansion}
	\end{minipage}
\end{figure}

\begin{figure}[t]
	\centering
	\begin{minipage}{0.45\textwidth}
		\centering
		\subfigure[]{
			\label{subfig:schematic_dimension_extension}
			\includegraphics[width=0.8\textwidth]{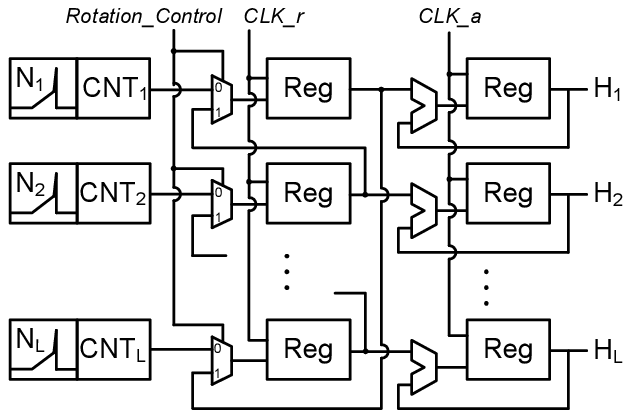}}
		\centering
		\subfigure[]{
			\label{subfig:timing_dimension_extension}
			\includegraphics[width=1\textwidth]{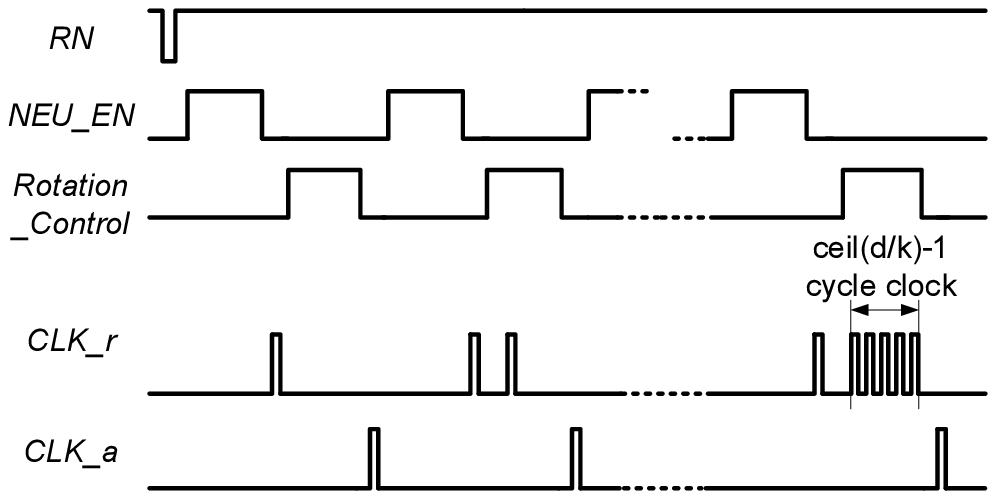}}
		\centering
		\caption{(a) Schematic of circuit for input dimension extension by shifting and summing the output counter values and (b) its timing diagram.}\label{fig:input_expansion}
	\end{minipage}
\end{figure}

\begin{figure}[t]
	\centerline{
		\includegraphics[width=0.2\textwidth]{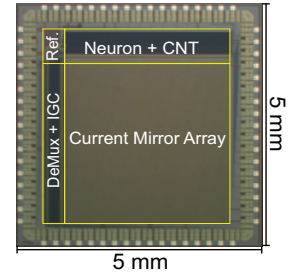} }
	\caption{Die photo of the prototype chip fabricated in $0.35 \mu$m CMOS.}\label{fig:die}
\end{figure}
\section{Input Dimension and Hidden Layer Extension Technique}
\label{sec:dimension}
For some applications, dimension of the input data is quite large (over several thousands) while other applications may require a large number of hidden layer neurons (also over several thousands) to achieve the best performance. This poses a big challenge to neuromorphic analog hardware implementations and have restricted the use of analog classifiers since the dimensions of the chip are fixed once fabricated. For example, suppose the input-dimension for an application is $d$ and it requires $L$ hidden layer neurons. Conventionally, at least $d\times L$ random weights are needed for the random projection operation in the first layer of ELM to get the hidden layer matrix $\bf{H}$. However if the maximum input dimension for the hardware is only $k$ ($k < d$) and the number of implemented hidden layer neurons is $N$ ($N < L$), the hardware can only provide a $k\times N$ random projection matrix $\bf{W}$ comprising weights $w_{ij}$(i = 1, 2, $\cdots$, k and j = 1, 2, $\cdots$, N). For more efficient use of the hardware, here we propose a method to reuse the input weights and hidden layer neurons to effectively expand both input dimension and number of hidden layer neurons beyond the number physically fabricated on-chip. Intuitively, each neuron requires $d$ random weights and there are a total of $k\times N$ such random weights on the chip. Hence, as long as $d<k\times N$, we can reuse these random weights to satisfy the requirement. Similarly, each input dimension requires $L$ random numbers for the projection--it can be attained by reusing weights as long as $L<k\times N$. A simple example of such an increased dimension of weight matrix is shown in Fig. \ref{fig:weight_reuse} for $k=2$ and $N=3$. This case shows the maximum dimension increase possible to get a matrix of size $(k\times N)\times (k\times N)$ Next, we elaborate the method used to do this assuming $d,L<k\times N$.

To expand the number of hidden layer neurons, we propose to do it in $\lceil{L/N}\rceil$ steps where the number of projections is increased $N$ in every step. For the second set of $N$ neurons, we need to shift the random matrix $\bf{W}$ comprising $w_{ij}$ (i = 1, 2,$\cdots$, d and j = 1, 2, $\cdots$, N) to $\bf{W}_{1,0}$ comprising $w_{ij}$ (i = 2, 3,$\cdots$, d, 1 and j = 1, 2, $\cdots$, N). Here, the subscript $(1,0)$ is used to denote a single circular rotation of the rows of the matrix $\bf{W}$. This notation implies $\bf{W}=\bf{W}_{0,0}=\bf{W}_{k,0}$. Using this notation, we can continue to get more random projections of the input (and thus expand the number of hidden neurons) by generating $\bf{W}_{1,0}$ to $\bf{W}_{\lceil{L/N}\rceil -1,0}$. Figure \ref{fig:hidden_expansion}(a) shows a simple circuit that can be added to the input side of the chip to achieve this function. The corresponding timing diagram of control signals are shown in Fig. \ref{fig:hidden_expansion}(b). Once the input data is loaded and the first set of hidden layer outputs are obtained (during the $NEU\_EN$ signal), the $Rotation\_Control$ signal is turned high to configure the input registers as a circular shift-register. This is followed by another $NEU\_EN$ signal to obtain the second set of $N$ random projections and this process continues till $L$ random projections are obtained.

A similar method can be applied to expand the input dimension from $k$ to $d$. In this case, we take the first $k$ dimensions $x_1,x_2..x_k$ of a particular input sample $x\in \Re^d$ and send it to the chip to get the multiplication for the first k dimensions with the random matrix $\bf W$. This generates $L$ hidden neuron outputs which can be expanded to a larger number using the technique described in the last paragraph. For the next $k$ dimensions of $x$, we shift the random matrix $\bf{W}$ comprising $w_{ij}$ (i = 1, 2, $\cdots$, k and j = 1, 2, $\cdots$, N) to $\bf{W}_{0,1}$ comprising $w_{ij}$ (i = 1, 2, $\cdots$, k and j = 2, 3, $\cdots$, N, 1). This implies a circular shift along the columns of $\bf{W}$. The hidden layer outputs obtained in this step are added to the ones obtained in the earlier step. This method can be continued for $\lceil{d/k}\rceil -1$ steps while accumulating the resulting hidden layer outs every time to get the final output for the $d$ dimensional input $x$.  Figure \ref{fig:input_expansion}(a) shows a simple circuit that can be added to the previously described chip architecture at the output to implement the input dimension expansion technique. Figure \ref{fig:input_expansion}(b) depicts the corresponding timing diagram. The circuit in Fig. \ref{fig:input_expansion}(a) shows a register bank after the neuron  output counters that can accept inputs from these counters or from other registers in this layer to effect the circular rotation of columns of $\bf{W}$. There is a second register bank after this which accumulates the counter outputs over multiple cycles. After the conversion of first $k$ dimensions of $x$ during the first $NEU\_EN$ signal, a clock pulse on $CLK\_r$ and $CLK\_a$ are used to shift this output to the accumulator. From the next cycle, the $Rotation\_Control$ signal is enabled and pulses on $CLK\_r$ are used to rotate the columns of the hidden layer. Another pulse on $CLK\_a$ is used to accumulate this value in the second register bank.

\begin{table}[htbp]
\centering
\caption{\label{summary} Chip Summary}
\begin{tabular}{|c|c|}
\hline
Technology & 0.35 $\mu m$ CMOS \\
\hline
Die Size & 5 mm $\times$ 5 mm \\
\hline
Input Channels  &  128 \\
\hline
Hidden Layer Size &  128 \\
\hline
Output Data format & 14-bit Digital  \\
\hline
Input Data format & 10-bit Digital \\
\hline
Power supply voltage & 1 V \\
\hline
\end{tabular}
\end{table}

  \begin{figure*} 
  	\begin{minipage}[b]{0.33\textwidth}
  		\centering
  		\includegraphics[width=1\textwidth]{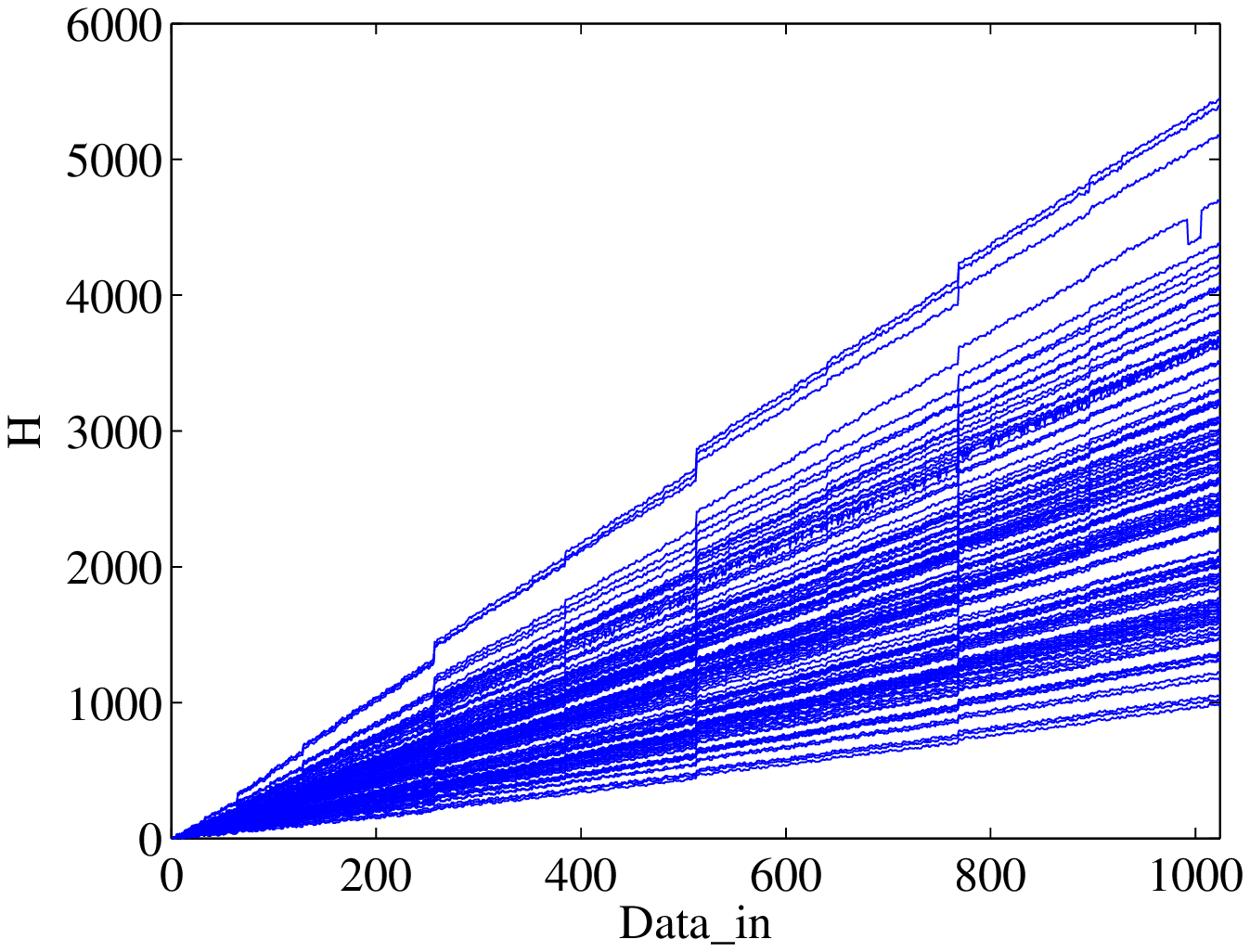} \\ (a)
  	\end{minipage}
  	\begin{minipage}[b]{0.33\textwidth}
  		\centering
  		\includegraphics[width=1\textwidth]{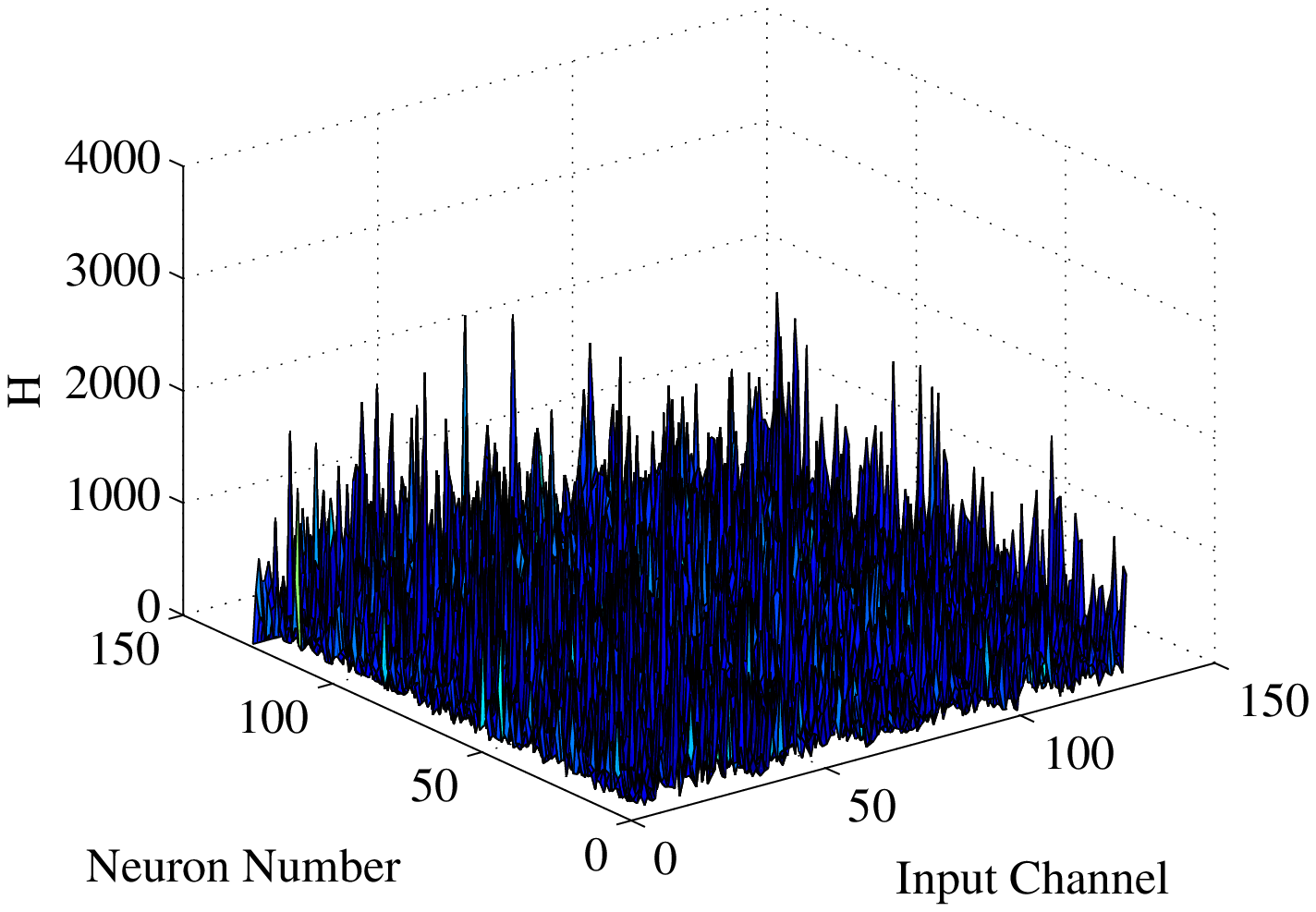} \\ (b)
  	\end{minipage}
  	\begin{minipage}[b]{0.33\textwidth}
  		\centering
  		\includegraphics[width=1\textwidth]{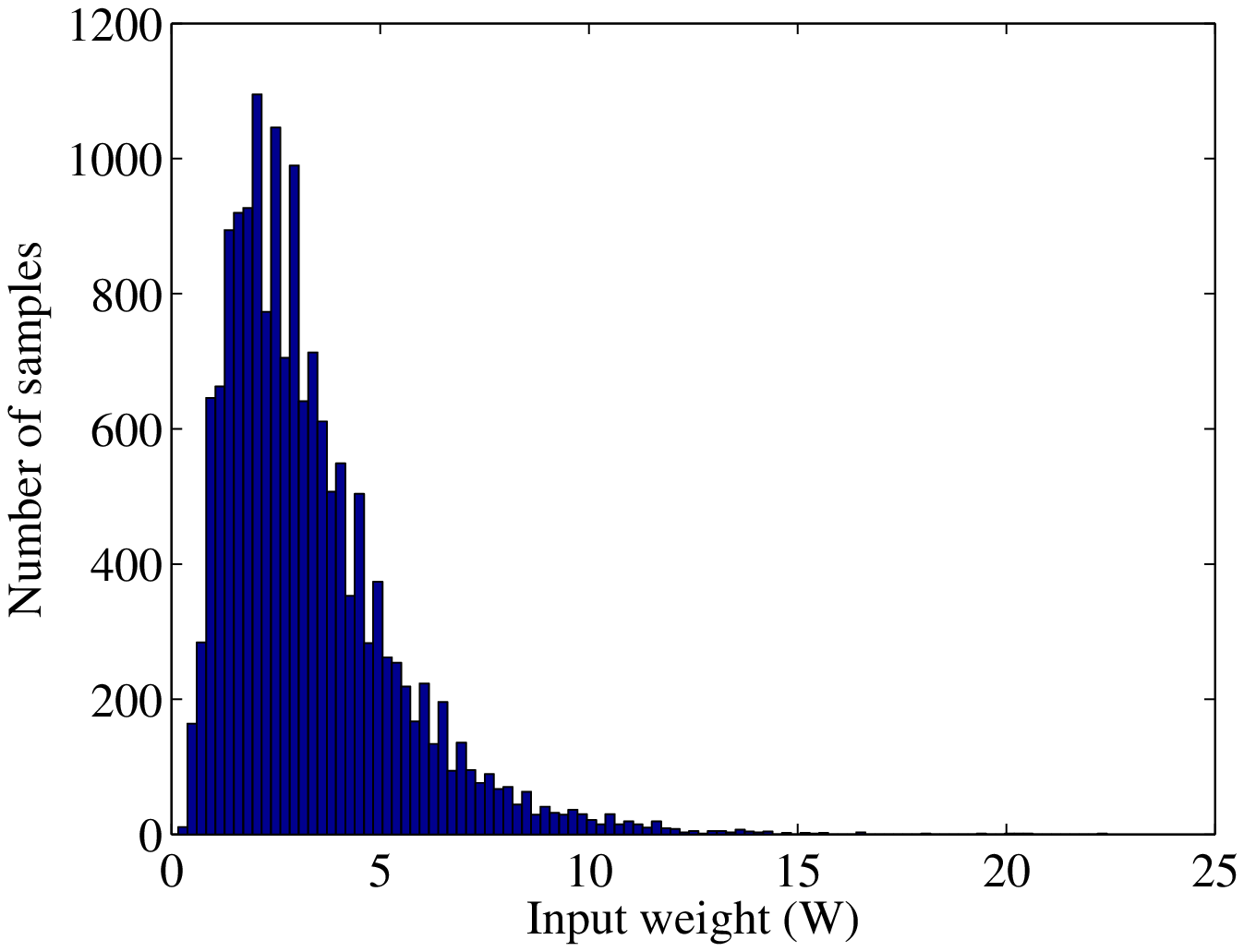} \\ (c)
  	\end{minipage}
  	\caption{ (a) Measured transfer function of hidden layer neurons when the digital input varying from $0$ to $1023$ with $d = 1$ and $T_{neu}=10$ms. (b) A surface plot showing the mismatch in weights of the $128\times 128$ current mirror synapses. The output counter values for different neurons are plotted for $T_{neu}=10$ms when $Data\_in=100$ is set on each input channel one by one. (c) Histogram showing the log-normal distribution of the input weights obtained from (b) for the $128 \times 128$ current mirror array.}
  	\label{fig:f_i_curve}
\end{figure*}

\begin{table*}[htbp]
	\centering
	\caption{\label{table:dataset} Measured performance on Binary Classification Datasets from UCI repository}
	\begin{tabular}{|c|c|c|c|c|c|}
		\hline
		\multirow{2}{*}{Datasets} & \multirow{2}{*}{\# Features ($d$)} & \multirow{2}{*}{\# Training} & \multirow{2}{*}{\# Testing} & \multicolumn{2}{c|}{Miss Classification Rate (\%)} \\
		\cline{5-6}
		& & & & {Software ($L=1000$) \cite{huang_elm_kernel}} &  {This work ($L=128$)} \\
		\hline
		Diabetes & 8 & 512 & 256 & 22.05  & 22.91 \\
		Australian Credit & 14 & 460 & 230 & 13.82  & 12.11\\
		Brightdata & 14 & 1000 & 1462 & 0.69  & 1.26 \\
		Adult & 123 & 4781 & 27780 & 15.41  & 15.57 \\
		\hline
	\end{tabular}
\end{table*}

\begin{figure}[t]
	\centerline{
		\includegraphics[width=0.4\textwidth]{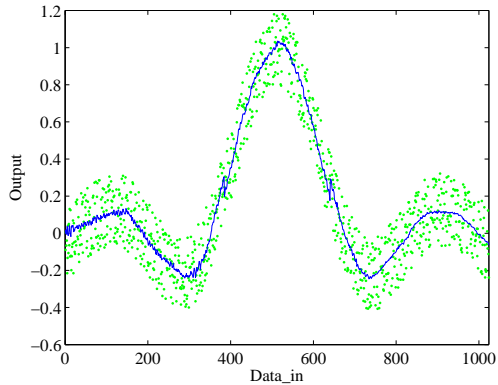} }
	\caption{Regression of underlying sinc function (in blue) based on a set of noisy samples (in green).}\label{fig:sinc_plot}
\end{figure}

\begin{table*}[htbp]
\centering
\begin{threeparttable}
\centering
\caption{\label{table:comparison} Comparison Table}
\begin{tabular}{|c|c|c|c|c|c|}
\hline
 & JSSC 2013 \cite{Verma_jssc2013} & JSSC 2007 \cite{shantanu_mvm} & IJCNN 2015 \cite{ijcnn_thakur} & ISCAS 2015 \cite{chenyi_iscas2015} & This work   \\
\hline
Technology & 0.13 $\mu$m & 0.5 $\mu$m  & 65 nm & 0.35 $\mu$m & 0.35 $\mu$m \\
\hline
Algorithm & SVM & SVM & ELM & ELM & ELM \\
\hline
Task & Classification & Classification & Regression & Regression & Regression \\
& & & & Classification & Classification \\
\hline
Design Style & Digital & Analog & Mixed mode & Mixed mode & Mixed mode \\
 & & Floating gate & & & \\
\hline
Supply Voltage & 0.85 V & 4 V & 1.2 V & 0.6 V (Digital) & 1 V \\
 & & & & 1.2 V (Analog) &  \\
\hline
Power Dissipation & 136.5 $\mu$W & 0.84 $\mu$W & - &  0.4 $\mu$W & 188.8 $\mu$W\tnote{$^1$} \\
\hline
Max Input Dimension & 400 & 14 & 1 & 128 & 16384\tnote{$^2$} \\
\hline
Energy Efficiency & 631 pJ/MAC\tnote{$^3$} & 0.8 pJ/MAC & - & 3.4 pJ/MAC\tnote{$^4$} & 0.47/ 0.54 pJ/MAC\tnote{$^5$} \\
\hline
Resolution & 16 b & 4.5 b & 13 b & 14 b & 14 b  \\
\hline
Classification Rate & 0.5-2 Hz & 40 Hz & - & 50 Hz & 31.6 kHz \\
\hline
Throughput & 2 MMAC/s & 1300 MMAC/s & - & 0.12 MMAC/s & 404.5 MMAC/s \\   
\hline
\end{tabular} 
\begin{tablenotes}
\item{$^1$} This power dissipation is measured based on $d=128$ and $L=100$.
\item{$^2$} Using input dimension extension technique to expand to $d=128\times 128$. Note that the circuits for rotating inputs and outputs for dimension increase are not included on this test chip.
\item{$^3$} Assuming $1000$ support vectors.
\item{$^4$} Only considering first stage of ELM for $d=40$ and $L=60$.
\item{$^5$} $0.47$ pJ/MAC is energy efficiency of current chip implementing first stage of ELM. The total energy per operation for binary classification is $0.54$ pJ/MAC using $VDD=1.5$ V for digital multipliers of second stage (see section \ref{sec:measure-power} for details). 
\end{tablenotes}
\end{threeparttable}
\end{table*}

\section{Measurement Results}
\label{sec:results}
\subsection{Characterization}
To validate the function of the proposed design, we have implemented the system in a $0.35\mu$m CMOS process. The ELM chip occupies a die area of $5mm\times 5mm$ as shown in Fig. \ref{fig:die}. The current area of the chip is dominated by the current mirror array since the layout is not optimized. Each cell in the current mirror array is pitch matched to the neuron in one direction and the IGC along another making it mostly empty. The area of the current mirror array can be reduced tremendously by following the proposal in \cite{yi-elm-decoder} limiting the size to the pitch of the IGC. In the next version, we will reduce the pitch of the IGC by moving to a scaled process like $65nm$. The mixed-signal chip implements the computationally intensive first stage while the second stage is currently implemented off-chip on a FPGA. In future, the second stage will also be integrated on the same die. Again, moving to a scaled process like $65$nm enables a small layout for this digital part. The larger statistical variation in a scaled process does not hurt the performance of the analog part as shown in Fig. \ref{fig:param_range}. The extra gate leakage in the current mirrors can be handled by either using thick oxide I/O devices or using active mirrors. Next, we present some characterization results to show the functionality of the chip. In all the experiments, both analog and digital power supplies are shorted together and is denoted by VDD. Unless stated otherwise, the default value of VDD$=1$V is used in most experiments.

First, we can get the transfer function of the $128$ neurons by sweeping the digital input $Data\_in$ on any one channel from $0$ to $1023$. The resultant curves are shown in Fig. \ref{fig:f_i_curve}(a). It can be seen that there is significant variation between the transfer curves of the neurons. Next, to characterize the random variation of the input weight matrix, we send a fixed value of $Data\_in$ to each of the input channels one by one and measure the counter outputs $H$. For every input channel, we get $L=128$ counter values indicative of the mismatch in that row. In total, there are $128\times 128$ such values of $H$ for all the input channels. These results are shown as a 3-dimensional plot in Fig. \ref{fig:f_i_curve}(b) where $H$ is plotted on the Z-axis. These same values are normalized by the median count value to get the effective weight distribution. This distribution of $128\times 128$ values is plotted as a histogram in Fig. \ref{fig:f_i_curve}(c) displaying a log-normal distribution. This is to be expected since $\Delta V_{Tn}$ has a normal distribution as explained in Section \ref{sec:mirror}. Further, by fitting a gaussian distribution to the logarithm of the weight values, we obtain $\sigma \Delta V_{Tn}\approx 16$mV in this process. Note that the mismatch obtained here also takes into account mismatch in the neuronal tuning curves since the count values are obtained at the output of the neuron. Further, this characterization is consistent across a set of $9$ chips with minimum and maximum values of $\sigma \Delta V_{Tn}$ being $15.36$mV and $16.26$mV respectively.

\subsection{Speed and Power}
\label{sec:measure-power}
During measurement, we found the chip to be functional for VDD down to $0.7$ V. Thus we can apply the results of the design space exploration in Section \ref{sec:energy} to optimize the system for the best speed and power efficiency. During measurement, a pico-ammeter (Keithley 6485) is utilized to measure the average current from the power supply to estimate the power dissipation. For all the experiments, speed and power are measured for $Data\_in=1000$ and $d=128$ with $L=100$ neurons activated. Conversion times $T_{neu}$ are estimated for $2^b=128$. At VDD$=0.7$V, the power dissipation is $17.85\mu$W at a maximum conversion speed of $4.5$kHz. As can be expected from Fig. \ref{fig:Ec_Imax}, there is not much variation in energy per classification when $I^z_{max}$ is reduced. However, this difference is more obvious at a higher VDD of $1$V. In this case, the fastest classification rate for this system is $146.25$ kHz corresponding to $T_{neu}=68.5\mu$s when $I^z\approx I_{flx}$. However, the power dissipation at this speed is quite high--$2.2$mW. Hence for a better energy efficiency, we optimize the classification rate to be around $31.6$ kHz by reducing $I^z_{max}$ to reduce the short-circuit power dissipation on $V_{mem}$ (as described in Section \ref{sec:energy- model}). The measured power dissipation now becomes $188.8 \mu$W as shown in Table \ref{table:comparison}. We choose this operating point as a good trade-off between speed and power efficiency. From this, we can approximate the coefficients $\alpha_1\approx 0.3 pF$ and $\alpha_2I_{sc}\approx 0.076 \mu A$ that are close to simulation values reported in section \ref{sec:energy- model}. Also, the analog power $P_{avdd}\approx 3.4 \mu W$. Considering the $128\times 100$ multiplication-and-accumulation (MAC) operation for the first layer, we can calculate the energy efficiency for this case as $0.47$ pJ/MAC. The corresponding throughput for classification rate of $31.6$ kHz is $404.5$ MMAC/s. Note that the current test chip does not have the digital multiplier for the second stage. Hence to estimate total system power, we have simulated a $14$-bit$\times$$10$-bit array multiplier in the same $0.35\mu$m process (assuming $b=14$ and resolution of $\beta=10$). For a digital $VDD=1.5$V, the energy per multiply is estimated to be $7.1$pJ at a delay of $12$ns. Using this value, the energy efficiency of the whole system for binary classification can be found to be $\approx 0.54$pJ/MAC.

\subsection{Regression and Classification}
\label{sec:tasks}
In order to verify the performance of the proposed neuromorphic ELM system in machine learning applications, we first show an example of regression ($d=1$) where the system was trained on $5000$ noisy samples (additive gausian noise with $\sigma = 0.2$) of a target $sinc(x)$ function and its task was to approximate the underlying function through regression. The input data is passed through the chip and hidden layer activations are obtained. These are next used for training the output weights. This method takes care of the mismatch in the neuronal transfer curves (which is also log-normal due to sub-threshold operation) by lumping it with the current mirror mismatch and training weights that take this into account. The measured result of this experiment are shown in Fig. \ref{fig:sinc_plot} for $L=128$ hidden neurons where the noisy samples are shown in green and the regressed function is in blue. The error of $0.021$ we obtain in this experiment is comparable to the error of $0.01$ obtained in software simulations of ELM\cite{huang_elm}. 

Next, we employ some real-world benchmark binary classification data sets from the UCI machine learning repository\cite{uci}. The reason for choosing these data sets are that they have different characteristics in terms of data dimension $d$ and data set size in terms of number of samples: small size and low dimensions ($Pima$ $Indians$ $diabetes$, $Statlog$ $ Australian$ $credit$), large size and low dimensions ($Star/Galaxy-Bright$), large size and high dimensions ($Adult$). The details of the data sets are shown in Table \ref{table:dataset}. During measurements, the hidden layer matrix $\bf{H}$ is obtained by applying the training data to the chip one by one. The second layer weights are obtained offline using this $\bf{H}$ and then downloaded to the FPGA for testing. The accuracy obtained in measurements with $L=128$ hidden neurons is shown in table \ref{table:dataset} and is compared with software simulation results taken from \cite{huang_elm_kernel}. This table shows that the performance of our implemented hardware ELM is comparable with the software ELM with the differences possibly due to the larger number of sigmoidal neurons (as opposed to saturating linear neurons for this chip) used in \cite{huang_elm_kernel}.

\subsection{Dimension Increase With Weight Reuse Technique}
In order to evaluate the performance for the dimension extension technique, we first applied a very high dimensional dataset ($leukemia$) with $d=7129$. Sizes for the training and testing data are $38$ and $34$ respectively. During measurement, we obtain a miss-classification rate of $20.59\%$ with $L=128$ neurons, which is comparable with the error rate of $19.92\%$ obtained using the software ELM reported in \cite{huang_elm_kernel}. Next, we separately prove the concept of artificially increasing number of hidden layer neurons. The measured errors in table \ref{table:dataset} are close to optimal and do not reduce much with further increase in $L$. Hence, we instead take $L=16$ neurons and use weight reuse method to expand to $L=128$. For the dataset $diabetes$, the error for $L=16$ is $27.1\%$. This reduces to an error of $22.4\%$, comparable to that in table\ref{table:dataset}, when $L$ is increased to $128$ by weight reuse. 
Note that since our chip did not have the circuits described in Section \ref{sec:dimension} to perform on-chip dimension expansion, we shifted the input data before applying it to the chip. Also, the output data was shifted in the FPGA before accumulation.

\subsection{Comparison}
Our work is compared with other recently reported hardware machine learners in Table \ref{table:comparison}. Our design is the most power efficient machine learner reported so far due to the low power analog multiplications. The energy efficiency of commercial digital processors are saturating at $\approx 100$pJ/MAC\cite{bo_powerwall}. Even custom digital multipliers have energy efficiencies of $10-70$pJ/MAC \cite{elm-biocas,mult-1,mult-2}. This explains the higher energy requirement of \cite{Verma_jssc2013} in Table \ref{table:comparison}. \cite{shantanu_mvm} uses analog floating-gate based multipliers and can hence achieve low-power multiplication. However, our approach does not require high voltages for programming floating-gates and is also much more compact due to the use of only one transistor without capacitors in the multiplier cell. \cite{ijcnn_thakur} also uses random mismatch (and a systematic offset) in $65$nm CMOS to perform the calculations in the first stage of ELM. However, they only have a single dimensional input and only show regression. Moreover, they do not report any energy or speed metrics. Lastly, compared to \cite{chenyi_iscas2015} which also uses the same core circuit of current mirrors to perform ELM computations for neural decoding, the current work is more energy efficient due to the faster operation (as explained in section \ref{sec:energy- model}). Also, the current work shows a method of expanding input dimension to a maximum of $d=16,384$ while \cite{chenyi_iscas2015} could only support a maximum of $d=128$. 

 \begin{figure}[h]
 	\centering 	
 	\includegraphics[width=0.35\textwidth]{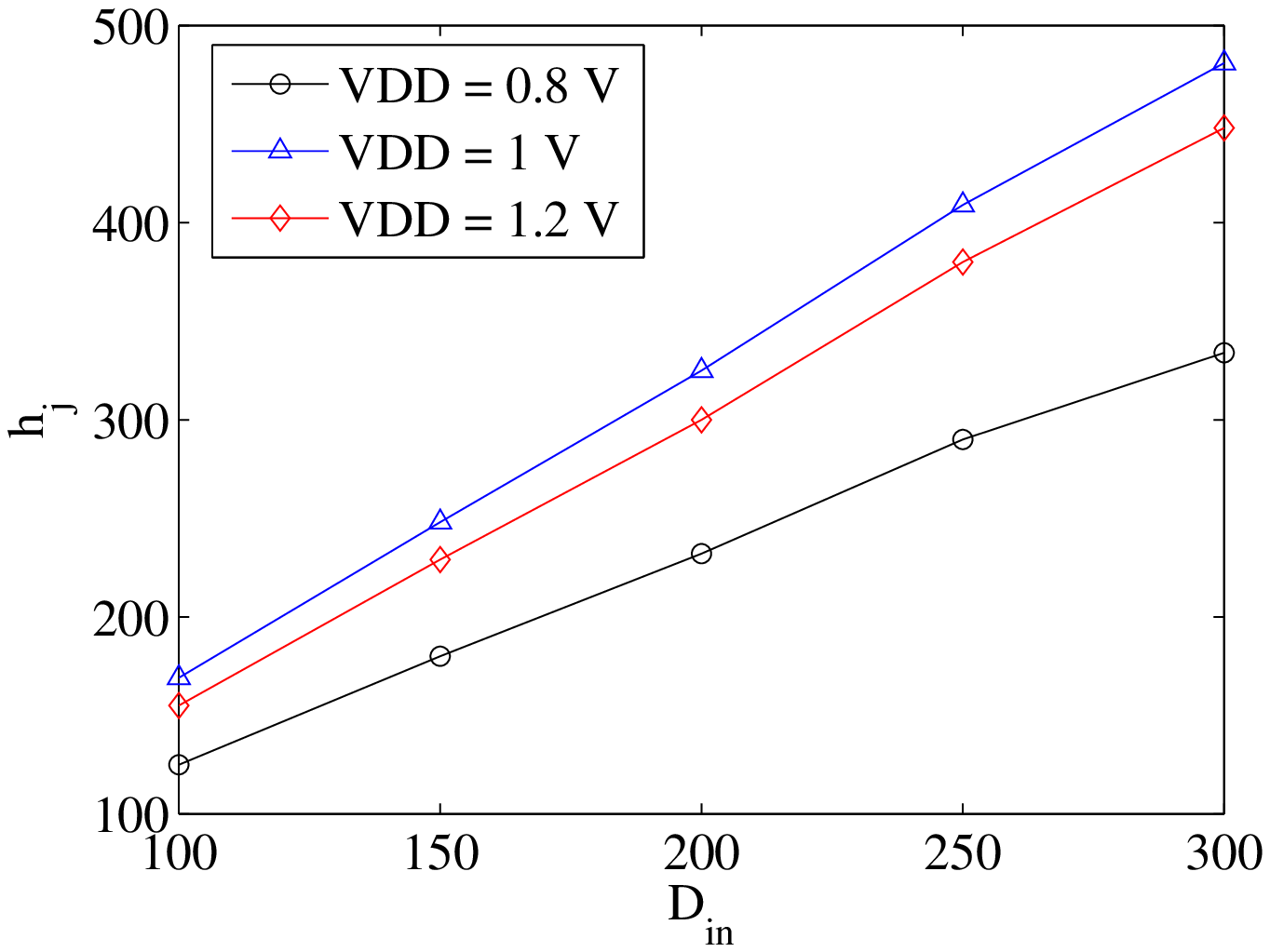} \\(a)\\
 	\includegraphics[width=0.35\textwidth]{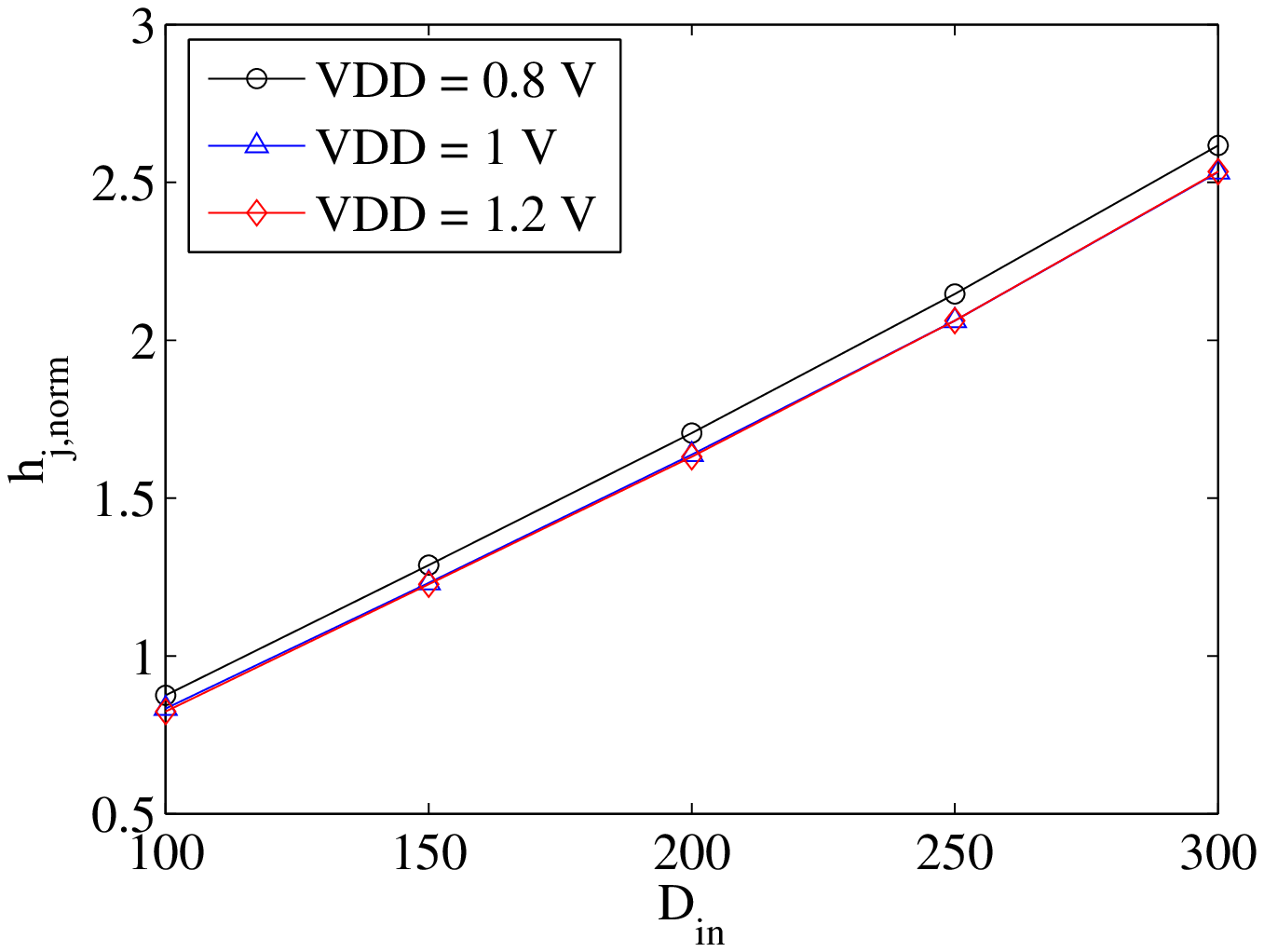} \\(b)
 	\caption{Comparison of hidden layer outputs for three different values of VDD in (a) the conventional case and (b) normalized case. The normalization results in less variation of output due to change in VDD.}\label{fig:vdd-vary}
 \end{figure}
 
 \begin{figure}[h]
  	\centering 	
  	\includegraphics[width=0.35\textwidth]{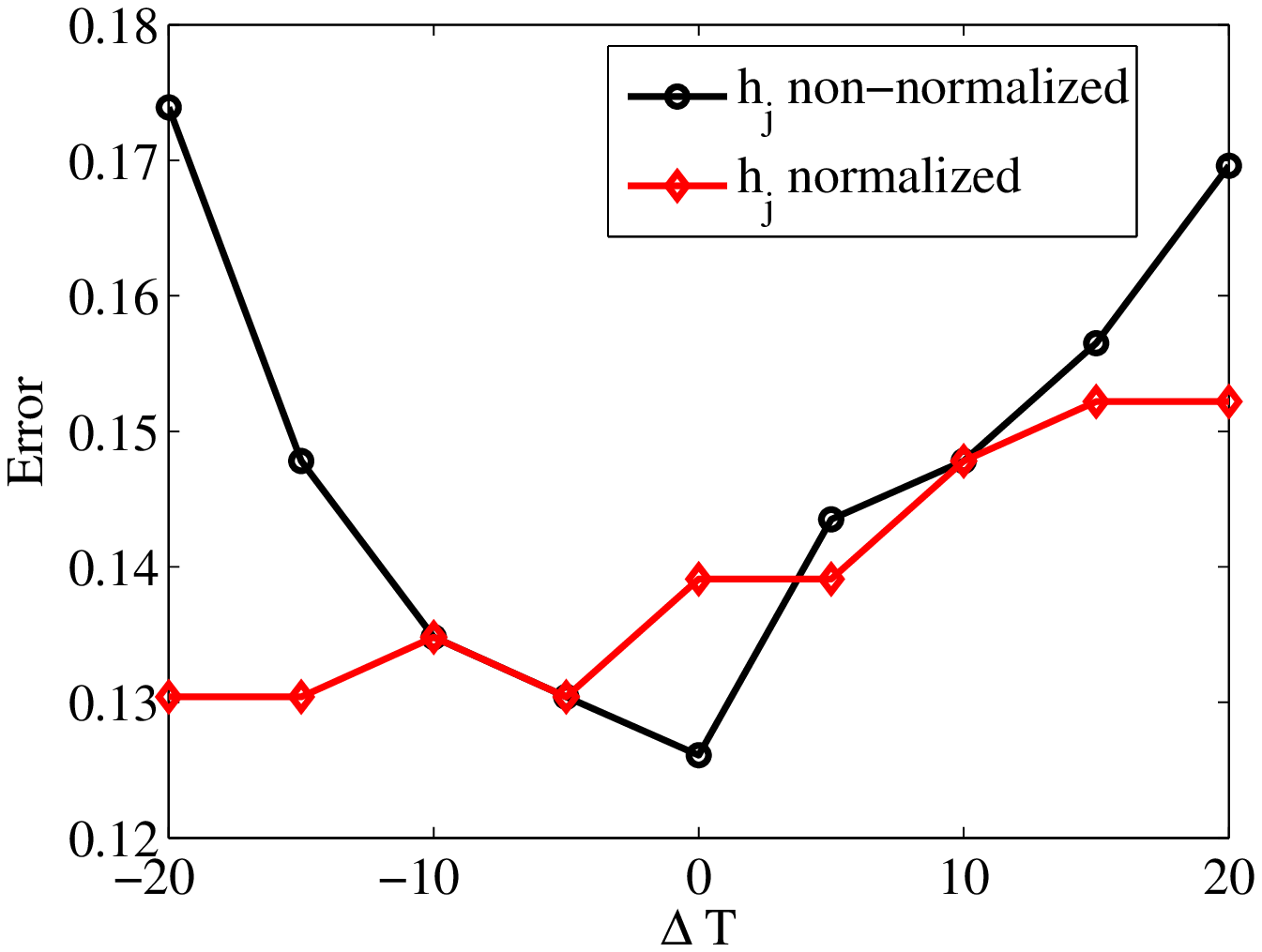} \\(a)\\
  	\includegraphics[width=0.35\textwidth]{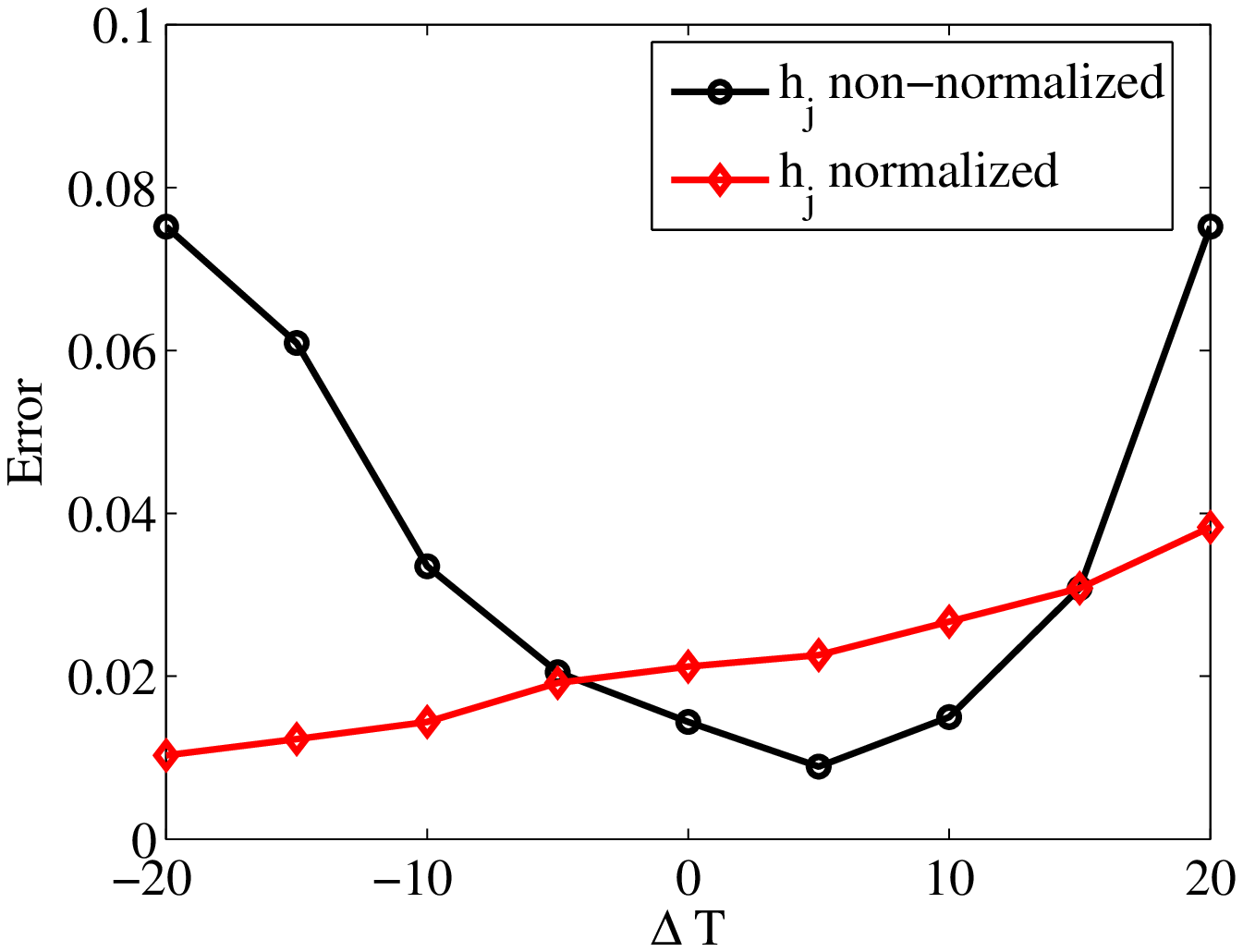} \\(b)
  	\caption{Comparison of performance when normalized and non-normalized hidden layer outputs are used for classification of (a) Australian credit and (b) Brightdata sets from the UCI repository.}\label{fig:temp-vary}
 \end{figure}

\begin{table}[t]
\centering
\caption{\label{table:variation_vdd} $Sinc$ function regression using normalized $h_j$}
\begin{tabular}{|c|c|c|}
\hline
\multirow{2}{*}{Power supply (V)} & Error (\%) & Error (\%) \\
 & (Non-normalized) & (Normalized) \\
\hline
0.8 & 0.5924 & 0.076 \\
\hline
1 & 0.045 & 0.0629 \\
\hline
1.2 & 0.1538 & 0.065 \\
\hline
\end{tabular}
\end{table}

\subsection{Robustness}
It is important to consider how the performance of the chip varies in the face of variations of power supply voltage (VDD) and temperature. We use the normalization method suggested in \cite{chenyi_iscas2015} to increase the robustness of our chip with respect to common-mode variations in VDD and temperature. Following, \cite{chenyi_iscas2015}, we define the j-th normalized hidden layer value ($h_{j,norm}$) as: 
\begin{align}
\label{eq:h-normalize}
h_{j,norm}&=\frac{h_j}{\sum_{j=1}^{L}h_j/\sum_{i=1}^{d}x_i}
\end{align}
To show the effectiveness of normalization, we first consider its effect on variations in VDD. Figure \ref{fig:vdd-vary}(a) plots measured values of hidden layer output $h_j$ for five different values of input data $D_{in}$ at three different values of VDD ($0.8$, $1$ and $1.2$V). It can be seen that there is a huge variation in $h_j$ (maximum of $22.7\%$). In contrast, when the same values are normalized (Fig. \ref{fig:vdd-vary}(b)), the variation due to change in VDD is reduced a lot (maximum of $4.2\%$) while variation due to change of $D_{in}$ is still retained. This proves effectiveness of the normalization method. We have further used the normalized and non-normalized values to perform the $sinc$ function regression task described in Section \ref{sec:tasks}. In this case, the weights are obtained for a nominal VDD of $1$V while testing is performed at all three VDD values. The result is reported in Table \ref{table:variation_vdd}. It can be seen that normalization enables the error to be low for all three values of VDD.

Next, we studied the effect of temperature variations on the hidden layer outputs. We expect the temperature dependent weights ($e^\frac{\Delta V_T}{U_T}$) to be the major contributor to variations in hidden layer outputs $h_j$. To confirm this prediction, we made a MATLAB model and obtained the variation of $h_j$ when temperature varied by $\Delta T=\pm 20\degree$C about a nominal value of $T_0=300$K. Then we benchmarked this variation with a SPICE simulation of the same circuit to confirm our earlier assumption--henceforth, we used the MATLAB model for simulations. Similar to the earlier case, we found that applying normalization reduced the maximum variation of hidden layer outputs from $9\%$ to $1.6\%$ over this temperature range. Next, we trained output weights for classification problems at the nominal temperature $T_0$ while the temperature was again varied over the same range during testing. We plot the results for $h_j$ and $h_{j,norm}$ for two different datasets in Fig. \ref{fig:temp-vary}(a) and (b). It can be seen that the error increases rapidly when temperature varies on either side of $T_0$ while using $h_j$. On the other hand, the error changes much more slowly when using $h_{j,norm}$ again confirming the benefit of normalization. Further, we have observed that retraining the weights can reduce the error close to the original value for both $h_j$ and $h_{j,norm}$. Hence, to get good performance over a wider range of temperature, we can store different weights for different tmperature ranges. One disadvantage with using the normalization is that now the second layer has to perform $L$ divisions on top of the $L\times C$ multiplications. But given the benefits provided, we believe that normalization is still a favourable choice. We do not have the normalization circuits included in this test chip but plan to include them in the next version.

\section{Conclusions}
\label{sec:conclusion}
We have presented a low-power hardware neuromorphic IC in $0.35\mu$m CMOS for machine learning applications using randomized neural networks such as random vector function link (RVFL), reservoir computing methods or extreme learning machines (ELM). Our hardware can also be used as a dimension reduction mechanism prior to applying unsupervised algorithms like k-nearest neighbors for clustering if the non-linear saturation in the neuron is not applied\cite{rp-knn-1,rp-knn-2}. The particular algorithm we employed in this work is extreme learning machine (ELM). The mismatch in silicon spiking neurons and synapses are used to perform the vector-matrix multiplication that forms the first stage of this classifier and is the most computationally intensive. Our results indicate that for a wide set of problems, $\sigma V_T$ in the range of $15-25$mV gives optimal results. A design space exploration is performed to show that minimum energy per operation at a specific VDD is obtained by operating for a short time at the highest spiking frequency achievable at that VDD. Linear neurons with a saturating non-linearity are used due to ease of implementation. Operating from a $1$ V power supply, this system can achieve an optimum energy efficiency of $0.47$ pJ/MAC with a corresponding classification rate of $31.6$ kHz making it one of the most energy efficient machine learners reported. Though this hardware can only implement randomized neural networks which might require a penalty of $2-3X$ more number of hidden nodes compared to networks with full tunability\cite{kitchen-sink} in many applications, the $10-20X$ lower energy required by random coefficient multiplications in our method overcome this penalty for lowering overall system energy. We also show a normalization method that enables a more robust operation of the circuit over changes in power supply and temperature. 

In future, we will apply this chip to classify multi-class image datasets such as MNIST. We will also explore the possibility of using it for dimension reduction prior to unsupervised clustering.


\begin{thebibliography}{10}

\bibitem{kinget-mismatch}
P.~R. Kinget,
\newblock ``{Device mismatch and tradeoffs in the design of analog circuits},''
\newblock {\em IEEE Journal of Solid-State Circuits}, vol. 40, no. 6, pp.
  1212--24, June 2005.

\bibitem{razavi-cmos}
B.~Razavi,
\newblock {\em {Design of Analog CMOS Integrated Circuits}},
\newblock Mc-Graw Hill Education, Aug 2000.

\bibitem{giacomo_mismatch2010}
E.~Neftci and G.~Indiveri,
\newblock ``{A device mismatch compensation method for VLSI neural networks},''
\newblock {\em IEEE Biomedical Circuits and Systems}, pp. 262--265, 2010.

\bibitem{giacomo_spikestdp}
G.~Indiveri, E.~Chicca, and R.~Douglas,
\newblock ``{A VLSI Array of Low-Power Spiking Neurons and Bistable Synapses
  With Spike-Timing Dependent Plasticity},''
\newblock {\em IEEE Transactions on Neural Networks}, vol. 17, no. 1, pp.
  211--221, Jan. 2006.

\bibitem{arthur_gamma}
J.~Arthur and K.~Boahen,
\newblock ``{Synchrony in Silicon: The Gamma Rhythm},''
\newblock {\em IEEE Transactions on Neural Networks}, vol. 18, no. 6, pp.
  1815--1825, Nov. 2007.

\bibitem{nullcline_neu_my}
A.~Basu and P.~Hasler,
\newblock ``{Nullcline based Design of a Silicon Neuron},''
\newblock {\em IEEE Transactions on Circuits and Systems I}, vol. 57, no. 11,
  pp. 2938--47, Nov. 2010.

\bibitem{bernabe-minidac}
B.~Linares-Barranco, T.~Serrano-Gotarredona, and R.~Serrano-Gotarredona,
\newblock ``{Compact low-power calibration mini-DACs for neural massive arrays
  with programmable weights},''
\newblock vol. 14, no. 5, pp. 1207--16, Sept 2003.

\bibitem{brink_learningfg}
S.~Brink, S.~Nease, and P.~Hasler et. al.,
\newblock ``{A Learning-enabled Neuron Array IC Based upon Transistor Channel
  Models of Biological Phenomenon},''
\newblock {\em IEEE Transactions on Biomedical Circuits and Systems}, vol. 7,
  no. 1, pp. 71--81, Feb. 2012.

\bibitem{sun-biocas}
S.~Shuo and A.~Basu,
\newblock ``{Analysis and reduction of mismatch in silicon neurons},''
\newblock in {\em IEEE Biomedical Circuits and Systems}, San-Diego, USA, Oct
  2011.

\bibitem{pfeil-stdp}
T.~Pfeil, A.~Scherzer, J.~Schemmel, and K.~Meier,
\newblock ``{Neuromorphic learning towards nano second precision},''
\newblock in {\em Proceedings of the International Joint Conference on Neural
  Networks}, Dallas, USA, 2013, pp. 1--5.

\bibitem{murray-mismatch}
K.~Cameron and A.~Murray,
\newblock ``{Can Spike Timing Dependent Plasticity compensate for process
  mismatch in neuromorphic analogue VLSI?},''
\newblock in {\em Proceedings of the International Symposium on Circuits and
  Systems}, Vancouver, 2004, pp. 748--51.

\bibitem{huang_elm_kernel}
G.-B. Huang, H.~Zhou, X.~Ding, and R.~Zhang,
\newblock ``{Extreme Learning Machine for Regression and Multiclass
  Classification},''
\newblock {\em IEEE Trans. on Systems, Man and Cybernetics- part B}, vol. 42,
  no. 2, pp. 515--29, 2012.

\bibitem{science-eliasmith}
Chris~Eliasmith et. al.,
\newblock ``{A large-scale model of the functioning brain},''
\newblock {\em Science}, vol. 338, no. 6111, pp. 1202--05, 2012.

\bibitem{skim}
J.~Tapson et. al.,
\newblock ``{Synthesis of neural networks for spatio-temporal spike pattern
  recognition and processing},''
\newblock {\em Frontiers in Neuroscience}, vol. 7, 2013.

\bibitem{elm-online}
A.~Van Schaik and J.~Tapson,
\newblock ``{Online and adaptive pseudoinverse solutions for ELM weights},''
\newblock {\em Neurocomputing}, vol. 149, pp. 233--8, 2015.

\bibitem{basu_shuo_elm}
A.~Basu, S.~Shuo, H.~Zhou, M.~H. Lim, and G.~B. Huang,
\newblock ``{Silicon Spiking Neurons for Hardware Implementation of Extreme
  Learning Machines},''
\newblock {\em Neurocomputing}, vol. 102, pp. 125--34, 2012.

\bibitem{elm-biocas}
Y.~Enyi, S.~Hussain, A.~Basu, and G.~B. Huang,
\newblock ``{Computation using Mismatch: Neuromorphic Extreme Learning
  Machines},''
\newblock in {\em Proceedings of the IEEE Biomedical Circuits and Systems
  Conference}, Oct 2013.

\bibitem{chenyi_iscas2015}
Yi~Chen, Enyi Yao, and Arindam Basu,
\newblock ``{A 128 channel 290 GMACs/W machine learning based co-processor for
  intention decoding in brain machine interfaces},''
\newblock in {\em Proceedings of the International Symposium on Circuits and
  Systems}, May 2015, pp. 3004--3007.

\bibitem{uci}
UCI Machine~Learning repository,
\newblock ``{http://archive.ics.uci.edu/ml/},'' .

\bibitem{huang_elm_survey}
G.-B. Huang, D.~H. Wang, and Y.~Lan,
\newblock ``{Extreme Learning Machines: A Survey},''
\newblock {\em Int. J. Mach. Learn. \& Cyber.}, vol. 2, pp. 107--122, 2011.

\bibitem{huang_elm}
G.-B. Huang, Q.~Y. Zhu, and C.~K. Siew,
\newblock ``{Extreme Learnng Machine: Theory and Applications},''
\newblock {\em Neurocomputing}, vol. 70, pp. 489--501, 2006.

\bibitem{1970_HoerlKennard}
Arthur~E. Hoerl and Robert~W. Kennard,
\newblock ``{Ridge Regression: Biased Estimation for Nonorthogonal Problems},''
\newblock {\em Technometrics}, vol. 12, no. 1, pp. 55--67, Feb. 1970.

\bibitem{tobi_bias}
T.~Delbruck and A.~Van Schaik,
\newblock ``{Bias current generators with wide dynamic range},''
\newblock {\em Analog Integrated Circuits and Signal Processing}, vol. 43, no.
  3, pp. 247--68, 2005.

\bibitem{giacomo_neuron_review}
G.~Indiveri et. al.,
\newblock ``{Neuromorphic Silicon Neuron Circuits},''
\newblock {\em Frontiers in Neuroscience}, vol. 5, no. 73, May 2011.

\bibitem{shantanu_mvm}
S.~Chakrabartty and G.~Cauwenberghs,
\newblock ``{A Sub-microwatt Analog VLSI Trainable Pattern Classifier},''
\newblock {\em IEEE Journal of Solid-State Circuits}, vol. 42, no. 5, pp.
  1169--1179, May 2007.

\bibitem{sarpeshkar}
R.~Sarpeshkar, T.Delbruck, and C.A. Mead,
\newblock ``{White noise in MOS transistors and resistors},''
\newblock {\em IEEE Transactions on Electron Devices}, vol. 9, no. 6, pp.
  23--29, Nov 1993.

\bibitem{Verma_jssc2013}
Kyong~Ho Lee and N.~Verma,
\newblock ``A low-power processor with configurable embedded machine-learning
  accelerators for high-order and adaptive analysis of medical-sensor
  signals,''
\newblock {\em IEEE Journal of Solid-State Circuits}, vol. 48, no. 7, pp.
  1625--1637, July 2013.

\bibitem{ijcnn_thakur}
C.~S. Thakur, T.~J. Hamilton, R.~Wang, J.~Tapson, and A.~V. Schaik,
\newblock ``{A neuromorphic hardware framework based on population coding},''
\newblock in {\em Proceedings of the International Joint Conference on Neural
  Networks}, Ireland, July 2015.

\bibitem{yi-elm-decoder}
Y.~Chen, Yao Enyi, and A.~Basu,
\newblock ``{A 128 channel Extreme Learning Machine based Neural Decoder for
  Brain Machine Interfaces},''
\newblock {\em IEEE Transactions on Biomedical Circuits and Systems}, 2015.

\bibitem{bo_powerwall}
B.~Marr, B.~Degnan, P.~E. Hasler, and D.~Anderson,
\newblock ``{Scaling Energy Per Operation via an Asynchronous Pipeline},''
\newblock {\em IEEE Transactions on VLSI}, vol. 21, no. 1, pp. 147--151, Jan.
  2013.

\bibitem{mult-1}
Y.~He and C.~H. Chang,
\newblock ``{A New Redundant Binary Booth Encoding for Fast 2n-Bit Multiplier
  Design},''
\newblock {\em IEEE Transactions on Circuits and Systems I}, vol. 56, no. 6,
  pp. 1192--1201, June 2009.

\bibitem{mult-2}
M.~La~Guia de~Solaz and R.~Conway,
\newblock ``{Razor Based Programmable Truncated Multiply and Accumulate,
  Energy-Reduction for Efficient Digital Signal Processing},''
\newblock {\em IEEE Transactions on VLSI}, vol. 23, no. 1, pp. 189--93, Jan
  2015.

\bibitem{rp-knn-1}
E.~Bingham and H.~Mannila,
\newblock ``{Random projection in dimensionality reduction: Applications to
  image and text data},''
\newblock in {\em Proceedings of the seventh ACM SIGKDD international
  conference on Knowledge discovery and data mining}, 2001, pp. 245--50.

\bibitem{rp-knn-2}
C.~Boutsidis, A.~Zouzias, and P.~Drineas,
\newblock ``{Random Projections for k-means Clustering},''
\newblock in {\em Proc. of Advances in Neural Information Processing Systems},
  2010, pp. 298--306.

\bibitem{kitchen-sink}
A.~Rahimi and B.~Recht,
\newblock ``{Weighted Sums of Random Kitchen Sinks: Replacing minimization with
  randomization in learning},''
\newblock in {\em Proc. of Advances in Neural Information Processing Systems},
  2009, pp. 1313--20.

\end{thebibliography}
\end{document}